\documentclass[12pt]{article}

\usepackage[margin=1in]{geometry}
\usepackage{amsmath}
\usepackage{amsfonts}
\usepackage{amssymb}

\usepackage{amsthm}
\usepackage[ruled,vlined]{algorithm2e}
\usepackage{mathtools}
\usepackage{tabularx}
\usepackage{graphicx}
\usepackage{subfigure}
\usepackage{enumerate}
\usepackage{float}
\usepackage{url}
\usepackage{verbatim}
\usepackage{cite}
\usepackage{hyperref}
\usepackage{makecell}
\usepackage{booktabs}
\usepackage{pifont}
\usepackage{multirow}
\usepackage{bm}
\usepackage[dvipsnames]{xcolor}
\usepackage[most]{tcolorbox}
\usepackage[inline]{enumitem}
\usepackage{indentfirst}
\usepackage{pdfpages}

\newtcolorbox{commentsbox}{breakable,enhanced,colframe=BrickRed,title=Comments}
\newtcolorbox{revisedbox}{breakable,enhanced,colframe=Blue,title=Revised,parbox=false}

\graphicspath{{figures/}}

\title{Supplementary Materials for \\
    \textbf{Collaborative Planning for Catching and Transporting Objects in Unstructured Environments}}

\author{Liuao Pei, Junxiao Lin, Zhichao Han, Lun Quan, \\Yanjun Cao, Chao Xu, and Fei Gao
	\thanks{All authors are with the State Key Laboratory of Industrial Control Technology, Zhejiang University, Hangzhou, 310027, China, and also with the Huzhou Institute of Zhejiang University, Huzhou, 313000, China. (\textit{Corresponding Author: Fei Gao})}
	\thanks{E-mail:{\tt\small \{plaa, fgaoaa\}@zju.edu.cn}}}
\begin{document}


# Collaborative Planning for Catching and Transporting Objects in Unstructured Environments

Liuao Pei, Junxiao Lin, Zhichao Han, Lun Quan, Yanjun Cao, Chao Xu, and Fei Gao

*Abstract*— Multi-robot teams have attracted attention from industry and academia for their ability to perform collaborative tasks in unstructured environments, such as wilderness rescue and collaborative transportation. In this paper, we propose a trajectory planning method for a non-holonomic robotic team with collaboration in unstructured environments. For the adaptive state collaboration of a robot team to catch and transport targets to be rescued using a net, we model the process of catching the falling target with a net in a continuous and differentiable form. This enables the robot team to fully exploit the kinematic potential, thereby adaptively catching the target in an appropriate state. Furthermore, the size safety and topological safety of the net, resulting from the collaborative support of the robots, are guaranteed through geometric constraints. We integrate our algorithm on a car-like robot team and test it in simulations and real-world experiments to validate our performance. Our method is compared to state-of-the-art multi-vehicle trajectory planning methods, demonstrating significant performance in efficiency and trajectory quality.

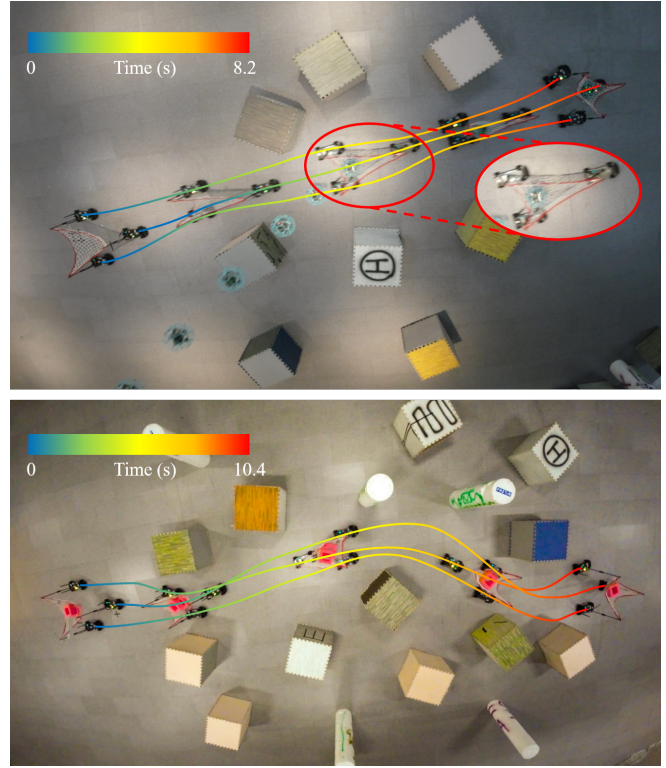

Fig. 1: Experimental results of a car-like robot team cooperatively catching and transporting objects in unstructured environments. Top: the robotic team collaboratively catches a falling drone with a supported net. Bottom: Some objects are transported through the unstructured environment by the robotic team.

## I. INTRODUCTION

With the advancement of autonomous navigation technology for car-like robots, some scenes depicted in science fiction movies, such as adaptive and flexible collaborative rescue and transportation in complex environments, have become a realistic possibility [1, 2]. The key to realizing these tasks lies in achieving precise and computing-efficient collaborative planning for the car-like robotic team, which is a high-dimensional, strongly-coupled system with non-holonomic motion constraints. However, achieving both high precision and high efficiency can be a dilemma, remaining challenging to perform online adaptive and precise collaborative trajectory planning in unstructured environments.

For collaborative catching and transportation tasks, in addition to considering individual motion constraints and obstacle avoidance, precisely and efficiently modeling the collaborative relationships among robotic teams is of paramount importance. When the robot team supports a net to perform catching and transporting tasks, there are multiple complex relationships among the robot team, the net, and the target to be rescued. Firstly, when a rescue target falls, the net supported by the robot team needs to catch the target in a theoretically feasible state of the robot team, defined as *adaptive catching*. The state of team formation when catching the target should be adjusted according to the environment and the distance to the target. Secondly, The trajectories need to keep the physical size of the net within the range of fatigue and damage at every moment, defined as *size safety*. At any given moment, the distance between any two robots must not exceed the corresponding distance limit associated with the two vertices. Thirdly, the net edges must not be entangled by any agent, defined as *topological safety*. From the initial time to any subsequent time, a robot should not pass through an edge of the net supported by any two other robots. The implementation of these collaborative relationships places great demands on agents in terms of both time and space, thereby creating a high-dimensional and strongly non-convex problem. However, there currently exists no method capable of planning the trajectories of non-holonomic robot teams to complete the catching and transporting tasks online.

To bridge this research gap, we propose an efficient centralized planning framework for car-like robotic teams.

Based on this framework, we model catching targets in a continuous and differentiable form for any feasible robot formation, allowing for fast gradient descent in optimization. As a result, the robot team is capable of fully exploiting its potential, deforming itself to a more appropriate state to catch the target. Furthermore, we impose restrictions on the size of the net and the topology of the team formation by accurately modeling the formation relationships. The *size safety* and *topological safety* of the rescue net supported by the robot team are guaranteed during navigation. Moreover, due to the adoption of efficient trajectory representation and decision variables transformation in optimization, all of these collaborative trajectories planning can be effectively completed online, making it possible for the robot team to perform catching and transporting tasks timely and accurately.

We apply our algorithm to a team of car-like robots and conduct catching and transporting experiments for falling targets in both simulation and real-world environments. As shown in Fig. 1, the experimental results demonstrate the ability of our algorithm to catch targets within its capability range in unstructured environments. As a foundational framework for collaborative planning, our proposed method is compared with some state-of-the-art multi-vehicle trajectory planning (MVTP) methods [3]–[7]. The results demonstrate that our method has significant advantages in terms of both computational time and success rates as agents number increases. Our contributions are summarized as follows:

1) We propose an efficient collaborative car-like robot team planning method that enables the generation of safe trajectories in unstructured environments with kinodynamic feasibility.
2) For real-time collaborative catching and transporting planning, we model multiple essential collaboration relationships and implement them based on the above framework.
3) We integrate our proposed method into a car-like robot team and validate the method in simulations and real-world experiments. We release our software for the community's reference.

## II. RELATED WORK

Currently, methods for solving the MVTP problem are mainly divided into coupled methods and decoupled methods. In order to reduce the enormous scale of the problem brought by coupled methods, the decoupled method decomposes the MVTP problem into several sub-problems and solves them gradually. Li et al. [8] proposed a prioritized trajectory optimization method for non-holonomic mobile robots. The method initially obtains collision-free paths using enhanced conflict-based search (ECBS) [9] and establishes safe corridors based on these paths. They improved the overall solution efficiency through grouping strategies. A penalty cost of deviating from the initial trajectory was introduced for the success rate of the optimization, which significantly limited the solution space. Moreover, sequential optimization can lead to deadlock issues in dense environments, resulting in failure to solve the problem. Luis et al. [7] proposed a distributed model predictive control (DMPC) based multi-agent motion planning framework and used on-demand collision avoidance for partitioning the free space, which allows for real-time trajectory planning with the less conservative movement and faster transition times. However, DMPC focused on planning in the range of specific horizon, which means the complete trajectories of the agents are often not optimal.

To account for the future states of multiple agents in the solution, the centralized methods solve for the trajectories of all agents in a single problem. Chen et al. [6] solved the MVTP problem by tightening collision constraints incrementally, thus forming a sequence of more relaxed and feasible intermediate optimization problems. They considered agents as circles for avoidance is conservative, and it may not guarantee that each quadratic programming (QP) problem is solvable when the kinematics are complex. Li et al. [10] proposed an adaptive-scaling constrained optimization (ASCO), which divides the intractably scaled constraints in the coupled optimal control problem (OCP) and conquers them in an iterative framework. In each iteration, partial collision avoidance constraints are implemented within an adaptive range. Ouyang et al. [5] proposed a two-stage method, which identifies a feasible homotopy class and finds a local optimum based on the identified homotopy class. However, they considered the car-like robot as two circles for obstacle avoidance, which is conservative, and the problem uses dense states as optimization variables, which makes it challenging to guarantee real-time performance. Wen et al. [11] proposed a spatiotemporal hybrid state A* for non-holonomic kinematic models. They introduced a hierarchical search-based solver called Car-like Conflict-Based Search (CL-CBS) and applied a body conflict tree to address collisions considering the shapes of the agents.

## III. COLLABORATIVE TRAJECTORIES OPTIMIZAITON

In this section, we introduce the motion model and differential flatness property of the car-like robots [12] and formulate collaborative planning as an optimization problem. Then we present the constraint handling approach, the basic kinematic constraints, and environmental obstacle avoidance constraints for the agents.

### A. Differential Flatness of the car-like robots

We use the simplified kinematic bicycle model in the Cartesian coordinate frame to describe the kinematics of car-like robots. Assuming that the car is front-wheel driven and steered with perfect rolling and no slipping, the model can be described as

$$\begin{cases} \dot{p}_x = v\cos\theta, \dot{p}_y = v\sin\theta, \\ \dot{\theta} = \frac{1}{L}v\tan\phi, \\ \dot{v} = a_t, \dot{a}_n = v^2\kappa, \\ \kappa = \tan\phi/L. \end{cases} \quad (1)$$

The state vector is $\boldsymbol{x} = (p_x, p_y, \theta, v, a_t, a_n, \phi, \kappa)^T$, where $\boldsymbol{p} = (p_x, p_y)^T$ denotes the position at the center of the rear

wheels, $\theta$ is the yaw angle of the robot's body, $v$ is the longitudinal velocity with respect to the robot's body frame, $a_t$ is the longitudinal acceleration, $a_n$ is latitude acceleration, $\phi$ is the steering angle of the front wheels, $\kappa$ is the curvature and $L$ is the wheelbase length of the car-like robot. We choose two differentially flat outputs $\boldsymbol{\sigma} := (\sigma_x, \sigma_y)^T$ as the optimization variables with a physical meaning that $\boldsymbol{\sigma} = \boldsymbol{p}$ is the position centered on the rear wheel of the car and the dynamics are given by:

$$\begin{cases} v = \eta\sqrt{\dot{\sigma}_x^2 + \dot{\sigma}_y^2}, \\ \theta = \arctan 2(\eta\dot{\sigma}_x, \eta\dot{\sigma}_y), \\ a_t = \eta(\dot{\sigma}_x\ddot{\sigma}_x + \dot{\sigma}_y\ddot{\sigma}_y)/\sqrt{\dot{\sigma}_x^2 + \dot{\sigma}_y^2}, \\ a_n = \eta(\dot{\sigma}_x\ddot{\sigma}_y - \dot{\sigma}_y\ddot{\sigma}_x)/\sqrt{\dot{\sigma}_x^2 + \dot{\sigma}_y^2}, \\ \phi = \arctan\left(\eta(\dot{\sigma}_x\ddot{\sigma}_y - \dot{\sigma}_y\ddot{\sigma}_x)L/(\dot{\sigma}_x^2 + \dot{\sigma}_y)^{\frac{3}{2}}\right), \\ \kappa = \eta(\dot{\sigma}_x\ddot{\sigma}_y - \dot{\sigma}_y\ddot{\sigma}_x)/(\dot{\sigma}_x^2 + \dot{\sigma}_y^2)^{\frac{3}{2}}, \end{cases} \quad (2)$$

where the additional variable $\eta$ is to characterize the motion direction of the robot with $\eta = -1$ and $\eta = 1$ representing backward and forward movements respectively [13].

### B. Optimization Problem Formulation

For the $k$-th agent in the swarm, the $i$-th segment of the trajectory is formulated as a 2-dimensional polynominal, which is divided into $M_{i,k}$ pieces with degree $2s - 1$, a coefficient matrix $\boldsymbol{c}_{i,k} = [\boldsymbol{c}_{i,1,k}^T, \boldsymbol{c}_{i,2,k}^T, ..., \boldsymbol{c}_{i,M_{i,k},k}^T]^T \in \mathbb{R}^{2sM_{i,k} \times 2}$ and a uniform time interval $T_{i,k} \in \mathbb{R}^+$. Then, this piece of the trajectory can be expressed as:

$$\begin{aligned} \boldsymbol{\sigma}_{i,j,k}(t) = \mathbf{c}_{i,j,k}^T \boldsymbol{\beta}(t), \quad \boldsymbol{\beta}(t) := [1, t, ..., t^{2s-1}]^T, \\ \forall t \in [0, T_{i,k}], \forall j \in \{1, 2, ..., M_{i,k}\}, \\ \forall i \in \{1, 2, ..., n_k\}, \\ \forall k \in \{1, 2, ..., K\}, \end{aligned} \quad (3)$$

where $K$ is the number of cooperative agents, $n_k$ is the number of trajectory of the $k$-th agent. The $i$-th segment of the trajectory $\boldsymbol{\sigma}_{i,k} : [0, T_{i,k}]$ is obteined:

$$\begin{aligned} \boldsymbol{\sigma}_{i,k}(t) = \boldsymbol{\sigma}_{i,j,k}(t - (j-1) * T_{i,k}), \\ \forall j \in \{1, 2, ..., M_{i,k}\}, t \in [(j-1) * T_{i,k}, j * T_{i,k}]. \end{aligned} \quad (4)$$

The total duration of the $i$-th segment of $k$-th agent's trajectory is $\tau_k = M_{i,k} * T_{i,k}$. Based on this, we can obtain the trajectory of the $k$-th agent:

$$\begin{aligned} \boldsymbol{\sigma}_k(t) = \boldsymbol{\sigma}_i(t - \tau_{i,k}), \\ \forall i \in \{1, 2, ..., n_k\}, t \in [\tau_{i,k}, \tau_{i+1,k}], \end{aligned} \quad (5)$$

where $\tau_k = \sum_{i=1}^{n_k} T_{i,k}$ is the duration of the whole trajectory, $\tau_{i,k} = \sum_{\bar{i}=1}^{i}$ is the timestamp of the starting point of the $i$-th segment and $\tau_1 = 0$. Moreover, we define a coefficient set $\mathbf{c}_k = [\mathbf{c}_{1,k}, \mathbf{c}_{2,k}, ..., \mathbf{c}_{n_k,k}]^T \in \mathbb{R}^{(\sum_{i=1}^{n_k} M_{i,k}s) \times 2}$ and time vector $\mathbf{T}_k = [T_{1,k}, T_{2,k}, ..., T_{n,k}]^T \in \mathbb{R}^n$ for $k$-th agent. Then we define the coefficient set $\mathbf{c} = [\mathbf{c}_1, \mathbf{c}_2, ..., \mathbf{c}_K] \in \mathbb{R}^{2M_is \times 2nK}$ and time vector $\mathbf{T} = [\mathbf{T}_1, \mathbf{T}_2, ..., \mathbf{T}_K]^T \in \mathbb{R}^{n_k}$ for the subsequent derivation.

The desired collaboration trajectories are optimized on the control effort, time regularization, and collaborative cost of all the agents to ensure smoothness, aggressiveness, and collaboration. Moreover, to guarantee that the trajectories are kinodynamic feasible and competent for the cooperation task, corresponding constraints are necessary. The optimization problem formulation is:

$$\min_{\mathbf{c}, \mathbf{T}} \sum_{k=1}^{K} \left( \int_0^{\tau_k} \boldsymbol{\mu}_k(t)^T \mathbf{W} \boldsymbol{\mu}_k(t) dt + w_T \tau_k \right) + \Psi_{obj}(\mathbf{c}, \mathbf{T}) \quad (6a)$$

$$\text{s.t.} \boldsymbol{\mu}_k(t) = \boldsymbol{\sigma}_k^{[s]}(t), \forall t \in [0, \tau_k], \quad (6b)$$

$$\boldsymbol{\sigma}_{i,k}^{[s-1]}(0) = \bar{\boldsymbol{\sigma}}_{0,i,k}, \quad (6c)$$

$$\boldsymbol{\sigma}_{i,j,k}^{[\widetilde{d}]}(\delta T_i) = \boldsymbol{\sigma}_{i,j+1,k}^{[\widetilde{d}]}(0), \quad (6d)$$

$$T_{i,k} > 0, \quad (6e)$$

$$\mathcal{G}_d(\boldsymbol{\sigma}(t), ..., \boldsymbol{\sigma}^{(s)}(t), t) \preceq 0, \forall d \in \mathcal{D}, \forall t \in [0, \tau_k], \quad (6f)$$

$$\forall i \in \{1, 2, ..., n_k\}, \forall j \in \{1, 2, ..., M_{i,k} - 1\},$$
$$\forall k \in \{1, 2, ..., K\},$$

where $\mathbf{W} \in \mathbb{R}^{2\times 2}$ is a diagonal matrix to penalize control efforts, $w_T \tau_k$ is the time regularization term, and $\Psi_{obj}(\mathbf{c}, \mathbf{T})$ is the collaborative cost when considering collaborative catching and transporting tasks, will be introduced in IV-C. Based on the proved optimality condition in the previous work [14], the entire piece-wise polynomial trajectory can be parameterized solely based on waypoints and the duration for each piece, thereby satisfying the equality constraints Eq.(6c-6d). For trajectory durations that are strictly constrained to be positive in Eq.(6e), we map them to unconstrained virtual times $\varsigma \in \mathbb{R}$ with a smooth bijection. Eq.(6f) is composed of environmental obstacle avoidance, inter-agent avoidance, kinodynamic constraints and collaborative constraints, where $\mathcal{D} = \{v, a_t, a_n, \kappa_l, \kappa_r, \mathfrak{C}, \Theta, \mathbb{L}_{net}, \mathbb{T}_{net}\}$ will be introduced in the following sections. We adopt the penalty term $S_\Sigma$ to relax the feasibility constraints Eq. (6f). Consequently, the original optimization problem is reformulated into an unconstrained formulation, which can be efficiently solved using the L-BFGS algorithm [15]. Further application details are introduced in Section 2 of the supplementary material [16].

### C. Single Agent Constraints

*1) Kinodynamic Constraints:* During the motion of the robot, the linear velocity, linear acceleration, latitude acceleration, and steering angle of the robot are limited by the physical properties of the robot motor, servo, tire, etc. These kinodynamic constraints can be expressed as follows:

$$\begin{cases} \mathcal{G}_v(\dot{\boldsymbol{\sigma}}) = \dot{\boldsymbol{\sigma}}^T\dot{\boldsymbol{\sigma}} - v_{max}^2, \\ \mathcal{G}_{a_t}(\dot{\boldsymbol{\sigma}}, \ddot{\boldsymbol{\sigma}}) = \dfrac{(\ddot{\boldsymbol{\sigma}}^T\dot{\boldsymbol{\sigma}})^2}{\dot{\boldsymbol{\sigma}}^T\dot{\boldsymbol{\sigma}}} - a_{tmax}^2, \\ \mathcal{G}_{a_n}(\dot{\boldsymbol{\sigma}}, \ddot{\boldsymbol{\sigma}}) = \dfrac{(\ddot{\boldsymbol{\sigma}}^T \mathbf{B}\dot{\boldsymbol{\sigma}})^2}{\dot{\boldsymbol{\sigma}}^T\dot{\boldsymbol{\sigma}}} - a_{nmax}^2, \\ \mathcal{G}_{\kappa_l}(\dot{\boldsymbol{\sigma}}, \ddot{\boldsymbol{\sigma}}) = \dfrac{\ddot{\boldsymbol{\sigma}}^T \mathbf{B}\dot{\boldsymbol{\sigma}}}{||\dot{\boldsymbol{\sigma}}||_2^3} - \kappa_{max}, \\ \mathcal{G}_{\kappa_r}(\dot{\boldsymbol{\sigma}}, \ddot{\boldsymbol{\sigma}}) = -\dfrac{\ddot{\boldsymbol{\sigma}}^T \mathbf{B}\dot{\boldsymbol{\sigma}}}{||\dot{\boldsymbol{\sigma}}||_2^3} - \kappa_{max}, \end{cases} \quad (7)$$

where $v_{max}$ is the magnitude of maximal logitude velocity, $a_{tmax}/a_{nmax}$ is the maximal longitude/latitude acceleration, $\mathbf{B} := [0\ -1;\ 1\ 0]$ is an auxiliary $2 \times 2$ matrix with the property of $\mathbf{B}^T = -\mathbf{B}$, $\kappa_{max}$ is the corresponding maximum curvature. Due to the monotonicity of tangent function, we restrict the front steer angle by limiting the curvature $\kappa = \tan\phi/L$ in $[-\tan\phi_{max}/L, \tan\phi_{max}/L] := [-\kappa_{max}, \kappa_{max}]$, where $\phi_{max}$ is the preset maximum steer angle.

*2) Environmental obstacle avoidance:* Considering the shape of the robot and in order to make full use of the safe solution space, we consider the geometric shapes for obstacle avoidance rather than as a point. After obtaining the grid map and the initial paths from hybrid A* [17], we sample the path to generate the safety corridor [18] and constrain the robot's whole body on the trajectory to be inside the safety corridor. We define the center of mass and orientation of the robot as the origin and x-direction of the body frame, which is illustrated in Fig.2. The rotation matrix $\mathbf{R}$, representing the orientation of the body frame relative to the world frame, can be expressed as:

$$\mathbf{R} = \frac{\eta}{\|\dot{\boldsymbol{\sigma}}\|_2}\left[\dot{\boldsymbol{\sigma}}, \mathbf{B}\dot{\boldsymbol{\sigma}}\right]. \quad (8)$$

We define the vertice set $\mathcal{E}$ of the $k$-th robot as:

$$\mathcal{E}_k = \{\boldsymbol{v}_e \in \mathbb{R}^2 : \boldsymbol{v}_e = \boldsymbol{\sigma} + \mathbf{R}\boldsymbol{l}_e, e = 1, 2, ..., n_e\}, \quad (9)$$

where $n_e$ is the number of vertexs and $\boldsymbol{l}_e$ is the coordinate of the $e$-th vertex in the body frame. The H-representation [19] of each convex polygon $\mathfrak{C}$ in the driving corridor is obtained:

$$\begin{aligned}\mathfrak{C} &= \{\boldsymbol{q} \in \mathbb{R}^2 : \boldsymbol{A}\boldsymbol{q} \leq \boldsymbol{b}\},\\ \boldsymbol{A} &= [\boldsymbol{A}_1, ..., \boldsymbol{A}_z, ..., \boldsymbol{A}_{n_z}]^T \in \mathbb{R}^{n_z \times 2},\\ \boldsymbol{b} &= [b_1, ..., b_z, ..., b_{n_z}]^T \in \mathbb{R}^{n_z},\end{aligned} \quad (10)$$

where $n_z$ is the number of hyperplanes, $\boldsymbol{A}_z \in \mathbb{R}^2$ and $b_z \in \mathbb{R}$ are the descriptors of the hyperplane, which can be determined by a point on the hyperplane and the normal vector. Therefore the constraint of the robot's whole body in the safety corridor on the trajectory can be expressed as:

$$\mathcal{G}_{\mathfrak{C}}(\boldsymbol{\sigma}, \dot{\boldsymbol{\sigma}}) = \boldsymbol{A}_z^T(\boldsymbol{\sigma} + \mathbf{R}\boldsymbol{l}_e) - b_z, \forall z \in \{1, 2, ..., n_z\} \quad (11)$$

## IV. COLLABORATIVELY CATCHING AND TRANSPORTING OBJECTS

### A. Inter-agents Obstacle Avoidance

To ensure the safety of agents, it is required that the convex multi-deformation distance under each moment between agents is greater than the safety distance $d_m$. Therefore, the constraint of collision avoidance between agents is defined as $\mathcal{G}_\Theta(\boldsymbol{\sigma}, \dot{\boldsymbol{\sigma}}) = [\mathcal{G}_{\Theta_1}, ..., \mathcal{G}_{\Theta_u}, ..., \mathcal{G}_{\Theta_{N_u}}]^T \in \mathbb{R}^{N_u}$ where $N_u = \frac{k(k-1)}{2}$ is the number of swarm agents in two-by-two combinations. The constraint of the $k$-th agent and $u$-th agent at a constraint point is defined as:

$$\mathcal{G}_{\Theta_{k,u}}(\boldsymbol{\sigma}_k, \dot{\boldsymbol{\sigma}}_k, \boldsymbol{\sigma}_u, \dot{\boldsymbol{\sigma}}_u) = d_m - U(\mathcal{E}_k, \mathcal{E}_u), \quad (12)$$

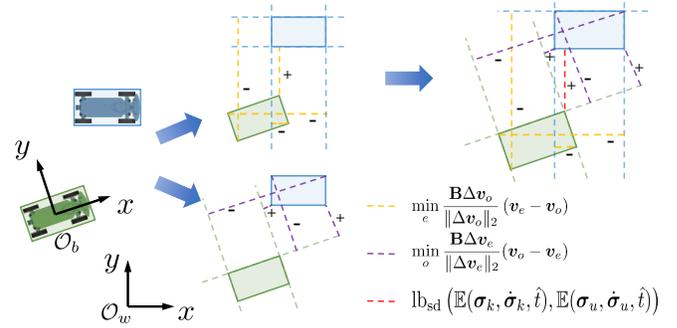

Fig. 2: The illustration of the lower bound of signed distance. First, we calculate the minimum distances from each vertex of one polygon to the edges of another polygon, resulting in purple and yellow dashed lines. Then, we determine the maximum value among these distances, which corresponds to the red dashed line. The positive or negative signs of the distances are indicated alongside them.

where $d_m$ is the minimum safe distance and $U(\mathbb{E}_k, \mathbb{E}_u)$ means the convex polygons distance between the $k$-th and $u$-th agent. In collaboratively trajectory optimization of agents, the computation of convex polygon distances between agents needs to be computed in a continuously differentiable manner. In this work, we use the maximum-minimum smoothing Minsky difference-based algorithm [20] to compute the lower approximation of the signed distances of convex polygons and obtain their derivatives, which can be expressed as:

$$\text{lb}_{\text{sd}}(\mathcal{E}_k, \mathcal{E}_u) = \max\left\{\max_e \min_o \frac{\mathbf{B}\Delta\boldsymbol{v}_e}{\|\Delta\boldsymbol{v}_e\|_2}(\boldsymbol{v}_o - \boldsymbol{v}_e),\right.$$
$$\left.\max_o \min_e \frac{\mathbf{B}\Delta\boldsymbol{v}_o}{\|\Delta\boldsymbol{v}_o\|_2}(\boldsymbol{v}_e - \boldsymbol{v}_o)\right\}, \quad (13)$$
$$\boldsymbol{v}_e \in \mathcal{E}_k, \boldsymbol{v}_o \in \mathcal{E}_u,$$

where $\Delta\boldsymbol{v}_e = \boldsymbol{v}_{e+1} - \boldsymbol{v}_e$ and $\Delta\boldsymbol{v}_u = \boldsymbol{v}_{u+1} - \boldsymbol{v}_u$. The physical meaning of the lower bound of the signed distance between two vehicles is shown in Fig. 2. To overcome the gradient discontinuity problem of eq.(13), we use the log-sum-exp function to smooth the maximum and minimum operations [21].

### B. Collaborative Transportation

In order to achieve the capability of cooperative transportation, usually the robot team cooperates to support a net, which requires maintaining the *size safety* and *topology safety* of the net. For *size safety*, we constrain the distance between each agent and the neighboring agents to be less than the corresponding net edge length, which can be expressed as:

$$\mathcal{G}_{\mathbb{L}_{net}}(\boldsymbol{\sigma}_k, \boldsymbol{\sigma}_{N(k)}) = \|\boldsymbol{\sigma}_k - \boldsymbol{\sigma}_{N(k)}\|_2 - L_k^{net}, \quad (14)$$
$$\forall k \in \{1, 2, ..., K\},$$

where $N(k)$ denotes the next agent number adjacent to the $k$-th agent, which can be expressed as:

$$N(k) = \begin{cases} k+1, & k < K, \\ 1, & k = K, \end{cases} \quad (15)$$

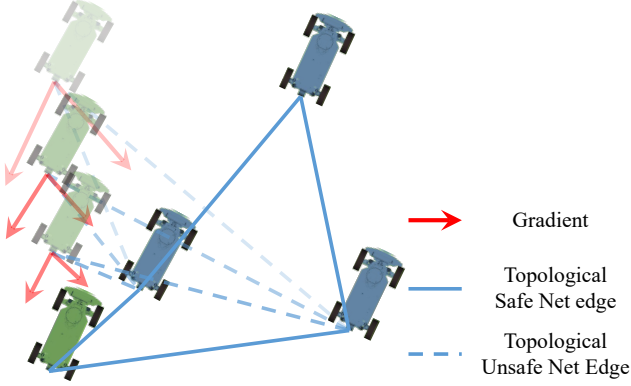

Fig. 3: The process of achieving topological safety during the optimization. The lightest shade of green represents the initial state of the green robot. By utilizing the aforementioned formulas, gradients in two directions are obtained and make the net achieve topological safety.

$L_k^{net}$ denotes the length of net edge between the $k$-th and $N(k)$-th agent. Moreover, in order to maintain the topology of the net during the collaborative support net, it is necessary to constrain the signed distance of the agent to the line where the non-neighboring net edges are located to be positive, which can be expressed as:

$$\mathcal{G}_{\mathbb{T}_{net}}(\boldsymbol{\sigma}_k, \boldsymbol{\sigma}_p, \boldsymbol{\sigma}_{N(p)}) = \frac{\mathbf{B}(\boldsymbol{\sigma}_{N(p)} - \boldsymbol{\sigma}_p)^T}{\|\boldsymbol{\sigma}_{N(p)} - \boldsymbol{\sigma}_p\|_2}(\boldsymbol{\sigma}_k - \boldsymbol{\sigma}_p),$$
$$\forall p \in \{p \in \{1, 2, ..., K\} : p \neq k, N(p) \neq k\}, \quad (16)$$
$$\forall k \in \{1, 2, ..., K\}.$$

The specific process is illustrated in Fig. 3. Through gradient descent, the collaborative robot team achieves topological safety. It is worth noting that the blue robot also adjusts its state simultaneously to ensure topological safety, although it is not depicted in the image for clarity.

By imposing two constraints on the positions of agents simultaneously along the trajectory, we ensure both *size safety* and *topological safety* of the net throughout the support period.

### C. Adaptive Catching

By coordinating the robot team to catch the object and solving it in the optimization, we can expand the successful feasible spatial-temporal solution space as much as possible in order to achieve catching at high-speed status in complex environments by agents.

For $E$ objects that need to be caught, sorted by the landing time, let $(\boldsymbol{p}_e, t_e)$ represent the predicted landing position and landing time, respectively. In order to ensure that the robot team is capable of completing the rescue mission in any formation, we utilize the following formula to describe the relationship between the landing point and the formation. Firstly, shoot a ray from the landing point in any direction, and check the number of intersection points between the ray and the polygon formed by the current formation (in accordance with IV.B). If it is odd, define the landing point as being inside the polygon with $\Upsilon = -1$. On the other

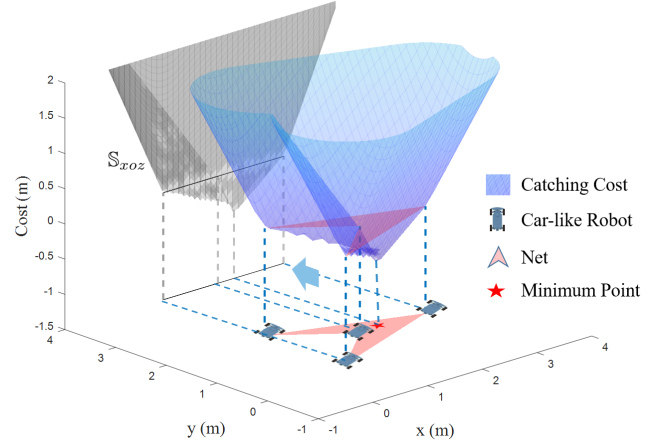

Fig. 4: The collaborative catching cost $\Psi_{obj}$ formed when the formation of the robot team is a non-convex polygon. During the optimization process, by changing the positions of each agent, the minimum point of the cost is made as close as possible to the landing point of the target.

hand, if it is even, define the landing point as being outside the polygon with $\Upsilon = 1$. Then, the signed distance between the point and the polygon formed by the formation can be expressed as:

$$\mathbb{D}(\boldsymbol{\sigma}, \boldsymbol{p}_e, t_e) = \Upsilon \min_k \left\{ \left\| -\boldsymbol{p}_e + \boldsymbol{\sigma}_k + \mathcal{L}\left(\frac{(\boldsymbol{\sigma}_{N(k)} - \boldsymbol{\sigma}_k)^T(\boldsymbol{p}_e - \boldsymbol{\sigma}_k)}{\|\boldsymbol{\sigma}_{N(k)} - \boldsymbol{\sigma}_k\|_2^2}\right)(\boldsymbol{\sigma}_{N(k)} - \boldsymbol{\sigma}_k) \right\|_2 \right\}, \quad (17)$$
$$k \in \{1, ..., K\}, e \in \{1, ..., E\},$$

where smoothing function $\mathcal{L}(x)$ is defined as:

$$\mathcal{L}(x) = \begin{cases} 0, & x \leq 0, \\ -\frac{1}{\epsilon^2}x^3 + \frac{2}{\epsilon}x^2, & 0 < x \leq \epsilon, \\ x, & \epsilon < x \leq 1 - \epsilon, \\ -\frac{1}{\epsilon^2}x^3 + \frac{3-2\epsilon}{\epsilon^2}x^2 \\ +\frac{4\epsilon-3}{\epsilon^2}x + \frac{(\epsilon-1)^2}{\epsilon^2}, & 1 - \epsilon < x \leq 1, \\ 1, & x > 1, \end{cases} \quad (18)$$

where $\epsilon \in \mathbb{R}^+$ is an excessive micro value. Based on this, we can calculate the signed distance between the target landing point and the formation of the robot team. We propose a collaborative cost function $\Psi$ for the collaborative catching of the target by the robots, expressed as:

$$\Psi_{obj} = \begin{cases} \omega_{ce}\mathbb{D}, & \mathbb{D} < -\epsilon, \\ \frac{\omega_{ce}-\omega_{ca}}{4\epsilon^3}\mathbb{D}^4 - \frac{3(\omega_{ce}-\omega_{ca})}{4\epsilon}\mathbb{D}^2 \\ +\frac{\omega_{ce}+\omega_{ca}}{2}\mathbb{D}, & -\epsilon < \mathbb{D} < \epsilon, \\ \omega_{ca}\mathbb{D}, & \mathbb{D} > \epsilon, \end{cases} \quad (19)$$

where $\omega_{ce}$ and $\omega_{ca}$ are the weights of taking the target to the center of the net and guaranteeing the catch of the target, respectively. As shown in Fig. 4, a non-convex polygonal formation can form a continuously differentiable cost based on Eq.(19), allowing the robot team to adaptively catch the

TABLE I: Comparison Of Trajectory Generation With State-of-the-art MVTP Methods

| Method | Success Rate | Time(s) | | | Average Travel Distance(m) | Average Acceration Cost($m^2s^{-3}$) | Global Planning | Centra- lization |
|---|---|---|---|---|---|---|---|---|
| | | Computation | Mean Travel | Longest Travel | | | | |
| Proposed, K = 8 | **100%** | 0.280 | **5.956** | 6.918 | 84.461 | 244.925 | | |
| Proposed, K = 14 | **100%** | **0.707** | 6.067 | **6.841** | **145.002** | 205.204 | | |
| Proposed, K = 20 | **100%** | 1.186 | 6.328 | 7.151 | 207.469 | 196.532 | **Yes** | **Yes** |
| Proposed, K = 26 | **100%** | 1.900 | 6.387 | 8.216 | 271.725 | 218.752 | | |
| FOTP [5], K = 8 | 100% | 4.897 | 6.169 | **6.169** | **81.462** | 286.648 | | |
| FOTP, K = 14 | 100% | 37.801 | 6.957 | 6.957 | 149.151 | 245.775 | | |
| FOTP, K = 20 | 100% | 147.214 | 7.835 | 7.835 | 222.223 | 204.270 | Yes | Yes |
| FOTP, K = 26 | 100% | 376.645 | 8.888 | 8.888 | 297.745 | **154.748** | | |
| MNHP [8], K = 8 | 85% | **0.077** | 5.009 | 5.009 | 84.779 | 957.611 | | |
| MNHP, K = 14 | 31% | - | 4.850 | 4.850 | 146.166 | 1462.799 | | |
| MNHP, K = 20 | 9% | - | 5.280 | 5.280 | 210.357 | 1705.211 | Yes | Hybrid |
| MNHP, K = 26 | 4% | - | 5.100 | 5.100 | 280.261 | 2701.124 | | |
| DMPC [7], K = 8 | 42% | 2.877 | 8.743 | 12.711 | 176.121 | 1213.8 | | |
| DMPC, K = 14 | 19% | 7.348 | 10.609 | 17.376 | 376.560 | 1597.272 | | |
| DMPC, K = 20 | 8% | 13.158 | 11.748 | 20.100 | 612.625 | 2043.097 | No | No |
| DMPC, K = 26 | 0% | - | - | - | - | - | | |
| SCP [22], K = 8 | **100%** | 164.020 | 5.000 | 5.000 | 79.480 | 272.010 | | |
| SCP, K = 14 | 92% | 1337.320 | 4.792 | 4.792 | 140.259 | 325.094 | | |
| SCP, K = 20 | 79% | 4065.292 | 4.884 | 4.884 | 202.888 | 322.018 | Yes | Yes |
| SCP, K = 26 | 15% | 9497.747 | 4.925 | 4.925 | 253.961 | 366.470 | | |
| CL-CBS [3], K = 8 | **100%** | 0.524 | - | - | 96.335 | - | | |
| CL-CBS, K = 14 | **100%** | 13.813 | - | - | 163.819 | - | | |
| CL-CBS, K = 20 | **100%** | 32.344 | - | - | 236.304 | - | Yes | Hybrid |
| CL-CBS, K = 26 | **100%** | 46.991 | - | - | 306.870 | - | | |

target and actively increase the formation when catching. In addition, in order to make the trajectory end more adaptable to the environment and smoother, we relaxed the end state of the trajectory.

$$\mathbf{M}_{i,k}(T_{i,k})c_{i,k} = \mathbf{b}_{i,k}, 1 \le i \le N_k, 1 \le k \le K, \quad (20)$$

where $\mathbf{M}_{i,k}$ is an invertible matrix defined in [14]. In this work, $\boldsymbol{b}_{n_k}$ is defined as follows:

$$\begin{aligned}\mathbf{b}_{n_k} = (&p_{n_k-1}, v_{n_k-1}, \mathbf{0}_{2\times(s-2)}, \boldsymbol{q}_{n_k,1}, \mathbf{0}_{2\times\tilde{d}}, ..., \\ &\boldsymbol{q}_{n_k,M_{n_k,k}-1}, \mathbf{0}_{2\times\tilde{d}}, \boldsymbol{\sigma}_k^f)^{\mathrm{T}}, \quad \forall k \in \{1, ..., K\},\end{aligned} \quad (21)$$

where $\boldsymbol{\sigma}_k^f$ is the final position of the $k$-th robot. We include the final positions of trajectories as decision variables. Then, the gradients of the objective function with respect to final positions are derived as follows:

$$\frac{\partial \mathcal{J}}{\partial \boldsymbol{\sigma}_k^f} = \left(\mathbf{M}_i^{-\mathrm{T}} \frac{\partial \mathcal{J}}{\partial c_i}\right)^{\mathrm{T}} e_{2sM_{n_k,k}}, \quad (22)$$

In this way, by optimizing the final positions of the trajectories, the robot team can freely choose a smoother parking position.

## V. EVALUATION AND EXPERIMENTS

### A. Benchmark Comparisons

To verify the effectiveness and computational efficiency of our proposed method, we compared it with several state-of-the-art MVTP methods. All the methods were tested on an Intel i7-12700 CPU with 32GB memory. We conducted tests on 100 cases with 8 random-sized and positioned obstacles. To ensure fairness in the comparative experiments, we provided the same initial guess for methods using the Hybird A* algorithm [17]. More details are introduced in Section 7 of the supplementary material [16]. We tested different numbers of agents and recorded success rates, computation time, the average time to execute agent trajectories, the average time to execute the longest trajectory, average total trajectory length, and average trajectory acceleration cost. These results are shown in Table I.

From the data, it can be seen that FOTP shows a high success rate, but due to the dense decision variables of all states and inputs in trajectories, the problem dimension is extremely high, leading to a huge computational burden. In contrast, our proposed method optimizes the waypoints and durations of each piece of the polynomial trajectory, resulting in a significant advantage in computation time compared to FOTP. On the other hand, our proposed method only needs to improve the initial path without considering collision avoidance to complete collision avoidance among agents in a relatively short time, significantly reducing the dependence

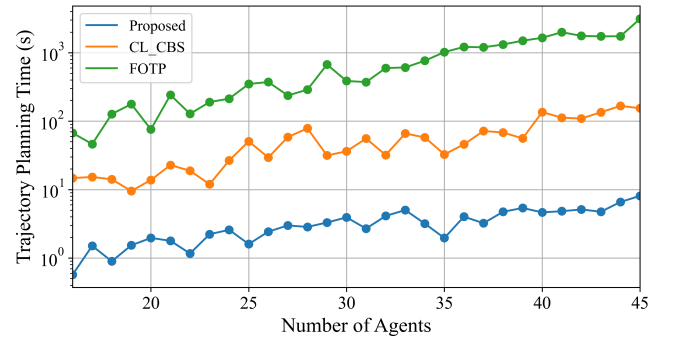

Fig. 5: Computation time under different agents numbers.

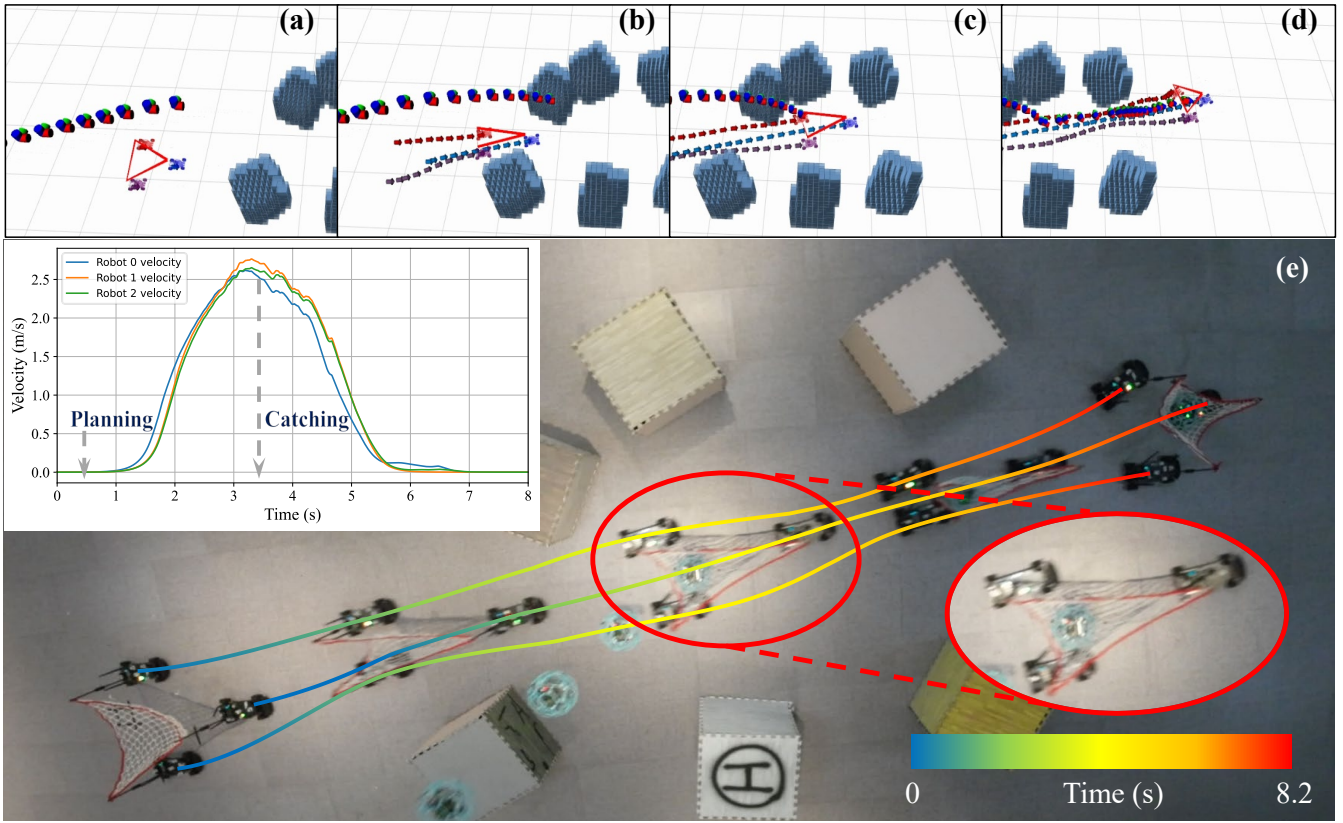

Fig. 6: Experimental validation of our method for a crashed aircraft rescuing. The collaborative rescue trajectory of the robot team is generated in 0.053 s and was able to rescue the downed aircraft successfully. (a) The initial state of the robot team and the aircraft began to experience a malfunction and consequently crashed. (b) The robot team transforms and maneuvers to avoid obstacles while maintaining the safety of agents and rescue net security. (c) The primary robot team proactively increases the net rescue size before and after rescuing the target to enhance the success rate and provide cushioning. (d) The robot team safely comes to a gentle stop in the secure area. (e) The global trajectory planning results.

on initial paths. The MNHP method uses a distributed approach with grouping and priority strategies. However, there are several limitations associated with it. Firstly, it considers robots as circular shapes for obstacle avoidance, which is conservative and leads to reduced solution space. This limitation hampers its effectiveness in environments where the robot's shape needs to be taken into consideration, especially in narrow spaces where the robot's shape becomes crucial. Secondly, the MNHP method penalizes the extent to which trajectories deviate from the initial values provided by ECBS [9] in the cost function. This heavy dependency on initial values significantly impacts its performance. Testing revealed that when the weight of this cost component was set to 0, MNHP faced difficulties in achieving collision avoidance among agents. Thirdly, MNHP does not optimize trajectory time and requires all agents to reach their specified final states simultaneously. This constraint further diminishes the possibility of successful spatiotemporal collision avoidance. These three limitations collectively restrict the effectiveness of MNHP. DMPC uses a priority-based strategy and employs on-demand obstacle avoidance strategy. However, it only considers local information in the near future for each planning iteration, which leads to overly greedy behavior and poor trajectory quality. It often takes detours in areas with dense obstacles and agents, and may even get stuck in infeasible situations due to its greediness. All these factors result in a low success rate and long travel distances for DMPC. In contrast, our coupled method provides higher solution quality and success rates, while still maintaining a significant advantage in computational efficiency compared to DMPC. From the data, it can be seen that CL-CBS has a high success rate, but it only provides a path with a time sequence that can serve as a feasible solution, without any effect on smoothness. Moreover, because of its semi-decentralized approach to seeking solutions, it has to face the problem of rapidly increasing computation time in the continuously shrinking solution space and the lower priority agents having the worse trajectory quality.

We conducted tests on two methods with the highest success rates in the table, FOTP and CL-CBS, under different cases of planning for the number of agents, as shown in Fig. 5. It can be observed that as the number of agents increases, the computation time of the proposed method increases gradually at a magnitude smaller than that of other methods. However, FOTP, which is also based on optimization, reaches a computation time of over 1000 $s$ after 35 agents.

## B. Real-World Experiments

The experimental field is a $6m \times 12m$ field with random obstacles and the robots collaboratively support a net. The processor of the central planning platform is an 8th Intel i5-8700 with 16 GB RAM, and the trajectories are sent to the swarm agents for tracking online. We fix the vertices of a net to the rear of each of the three car-like robots and carry out a collaborative transportation experiment on the field. After being given the target state, the central computing platform calculates the trajectories of the robotic team, which is then followed by the MPC controllers of each agent [23]. The MPC serves as a general trajectory tracker for controlling the experimental robot's spatiotemporal trajectory. It utilizes the same model as introduced in Eq.(1), with the input being the robot team trajectory generated by the proposed method. Further application details are introduced in Section 3 of the supplementary material [16]. The catching process is shown in Fig. 6, where a failed drone acts as the rescue target. During the flight, the drone suddenly malfunctioned and fell. Assuming that we obtain the expected location and time of the drone's fall, the system begins to plan collaborative rescue trajectories. Our method generates collaborative trajectories for three car-like robots in just 0.053s. Throughout the process, in addition to satisfying the kinematics and safety of each agent's movement, both the *size safety* and *topological safety* of the net are also guaranteed. The generated trajectories highly exploit the robots' kinematic performance to complete the rescue mission. As shown in Fig 6(e), to successfully achieve long-distance rescue, the robot team first accelerates at almost maximum acceleration and achieves the rescue at almost maximum speed. Then, due to the relaxation of the terminal state at the rear end, they smoothly stop in the safe area.

## VI. CONCLUSION

In this work, we introduce a collaborative robotic team planning system with capabilities of collaborative target catching and transporting. By modeling cooperative tasks, we complete the planning of a car-like robot team to collaboratively catch and transport objects. Comparison with current multiple MVTP methods shows the solution quality and scalability of our method. Simulations and real-world experiments demonstrate the robustness and flexibility of our method. In the future, one research direction is to extend our method to transportation in three-dimensional environments, where the team formation changes to enable the transported object to avoid obstacles considering height.

## REFERENCES

[1] J. Alonso-Mora, R. Knepper, R. Siegwart, and D. Rus, "Local Motion Planning for Collaborative Multi-Robot Manipulation of Deformable Objects," in *IEEE International Conference on Robotics and Automation (ICRA)*, May 2015, pp. 5495–5502.

[2] H. Zhang, H. Song, W. Liu, X. Sheng, Z. Xiong, and X. Zhu, "Hierarchical motion planning framework for cooperative transportation of multiple mobile manipulators," *arXiv preprint arXiv:2208.08054*, 2022.

[3] L. Wen, Y. Liu, and H. Li, "CL-MAPF: Multi-agent path finding for car-like robots with kinematic and spatiotemporal constraints," *Robotics and Autonomous Systems*, vol. 150, p. 103997, 2022.

[4] J. Li, M. Ran, and L. Xie, "Efficient Trajectory Planning for Multiple Non-Holonomic Mobile Robots via Prioritized Trajectory Optimization," *IEEE Robotics and Automation Letters*, vol. 6, no. 2, pp. 405–412, Apr. 2021.

[5] Y. Ouyang, B. Li, Y. Zhang, T. Acarman, Y. Guo, and T. Zhang, "Fast and Optimal Trajectory Planning for Multiple Vehicles in a Nonconvex and Cluttered Environment: Benchmarks, Methodology, and Experiments," in *International Conference on Robotics and Automation (ICRA)*, May 2022, pp. 10 746–10 752.

[6] Y. Chen, M. Cutler, and J. P. How, "Decoupled Multiagent Path Planning via Incremental Sequential Convex Programming," in *IEEE International Conference on Robotics and Automation (ICRA)*, May 2015, pp. 5954–5961.

[7] C. E. Luis, M. Vukosavljev, and A. P. Schoellig, "Online Trajectory Generation With Distributed Model Predictive Control for Multi-Robot Motion Planning," *IEEE Robotics and Automation Letters*, vol. 5, no. 2, pp. 604–611, Apr. 2020.

[8] J. Li, M. Ran, and L. Xie, "Efficient trajectory planning for multiple non-holonomic mobile robots via prioritized trajectory optimization," *IEEE Robotics and Automation Letters*, vol. 6, no. 2, pp. 405–412, 2021.

[9] M. Barer, G. Sharon, R. Stern, and A. Felner, "Suboptimal variants of the conflict-based search algorithm for the multi-agent pathfinding problem," in *Proceedings of the International Symposium on Combinatorial Search*, vol. 5, no. 1, 2014, pp. 19–27.

[10] B. Li, Y. Ouyang, Y. Zhang, T. Acarman, Q. Kong, and Z. Shao, "Optimal Cooperative Maneuver Planning for Multiple Nonholonomic Robots in a Tiny Environment via Adaptive-Scaling Constrained Optimization," *IEEE Robotics and Automation Letters*, vol. 6, no. 2, pp. 1511–1518, Apr. 2021.

[11] L. Wen, Z. Zhang, Z. Chen, X. Zhao, and Y. Liu, "CL-MAPF: Multi-Agent Path Finding for Car-Like Robots with Kinematic and Spatiotemporal Constraints," *Robotics and Autonomous Systems*, vol. 150, p. 103997, Apr. 2022.

[12] R. M. Murray, M. Rathinam, and W. Sluis, "Differential Flatness of Mechanical Control Systems: A Catalog of Prototype Systems," in *Proceedings of the 1995 ASME International Congress and Exposition*, 1995.

[13] Z. Han, Y. Wu, T. Li, L. Zhang, L. Pei, L. Xu, C. Li, C. Ma, C. Xu, S. Shen, and F. Gao, "An efficient spatial-temporal trajectory planner for autonomous vehicles in unstructured environments," 2023.

[14] Z. Wang, X. Zhou, C. Xu, and F. Gao, "Geometrically Constrained Trajectory Optimization for Multicopters," *IEEE Transactions on Robotics*, vol. 38, no. 5, pp. 3259–3278, Oct. 2022.

[15] D. C. Liu and J. Nocedal, "On the limited memory bfgs method for large scale optimization," *Mathematical programming*, vol. 45, no. 1-3, pp. 503–528, 1989.

[16] L. Pei, *Supplementary Materials for Collaborative Planning for Catching and Transporting Objects in Unstructured Environments*, Sept. 2023. [Online]. Available: https://doi.org/10.5281/zenodo.8376633

[17] D. A. Dolgov, S. Thrun, M. Montemerlo, and J. Diebel, "Path planning for autonomous vehicles in unknown semi-structured environments," *The International Journal of Robotics Research*, vol. 29, pp. 485–501, 2010.

[18] X. Zhong, Y. Wu, D. Wang, Q. Wang, C. Xu, and F. Gao, "Generating large convex polytopes directly on point clouds," *arXiv preprint arXiv:2010.08744*, 2020.

[19] D. Avis, K. Fukuda, and S. Picozzi, "On canonical representations of convex polyhedra," in *Mathematical software*. World Scientific, 2002, pp. 350–360.

[20] S. Cameron and R. K. Culley, "Determining the minimum translational distance between two convex polyhedra," *IEEE International Conference on Robotics and Automation (ICRA)*, vol. 3, pp. 591–596, 1986.

[21] I. Goodfellow, Y. Bengio, and A. Courville, *Deep learning*. MIT press, 2016.

[22] F. Augugliaro, A. P. Schoellig, and R. D'Andrea, "Generation of collision-free trajectories for a quadrocopter fleet: A sequential convex programming approach," *IEEE/RSJ International Conference on Intelligent Robots and Systems*, pp. 1917–1922, 2012.

[23] J. Kong, M. Pfeiffer, G. Schildbach, and F. Borrelli, "Kinematic and dynamic vehicle models for autonomous driving control design," *2015 IEEE Intelligent Vehicles Symposium (IV)*, pp. 1094–1099, 2015.


\date{}
\maketitle

\section{Modeling Process}\label{S1}

Here, we provide a rearticulation of the modeling process for this centralized planning.

To begin with, we formulate the problem as a nonlinear optimal control problem. However, due to the highly nonlinear nature and the high dimensionality of the optimization problem, in this work, we employ differential flatness for simplification without sacrificing optimality. This allows us to express the original robot state and control variables analytically using flat outputs and their finite-dimensional derivatives.

Subsequently, we parameterize the flat outputs using piece-wise polynomials, which naturally ensure motion continuity while reducing the dimensionality of the optimization, thereby enhancing the computational efficiency of the solution. Overall, the application of differential flatness aims to simplify the optimization problem and improve real-time computational efficiency. The specific formulation of the simplified problem can be found in \hyperref[sec:opt_solve]{Sec. 2}.

Furthermore, we instantiate the formulation of constraints $\mathcal{G}$ in the paper and analytically derive their gradients with respect to the optimization variables (detailed information can be found in our paper), enabling the efficient solution using the quasi-Newton method.

We hope that this clarification of the problem elucidates the framework and modeling process of our approach. Additionally, we have made corresponding revisions to the paper based on your feedback. Thank you again for your valuable suggestions.

\section{Implementation Details on Optimization Problem Solving}\label{sec:opt_solve}

Intuitively, we analyze the characteristics of each constraint and subsequently eliminate them individually. As a result, the initial optimization problem is restructured into an unconstrained optimization, allowing us to employ a robust and commonly used quasi-newton method for its solution. Here, we give a detailed introduction to the implementation details.

First, we give the initial formulation of the optimization problem:
\begin{subequations}
	\begin{align}
		\min_{\bm{{\rm c}},\bm{{\rm T}}}  &\sum\limits_{k=1}^{K}\! \left(\! \int_{0}^{\tau_{k}}\! \bm{\mu}_{k}(t)^T \bm{ {\rm W} } \bm{\mu}_{k}(t) dt\! +\! w_T \tau_{k}\!\right)\! +\! \Psi_{obj}(\bm{{\rm c}},\!\bm{{\rm T}})\!  \\
		{\rm s.t.} &\bm{\mu}_{k}(t) = \bm{\sigma}_{k}^{[s]}(t),  \forall t \in [0, \tau_{k}],   \quad \quad \quad\\
		&\bm{\sigma}^{[s-1]}_{i,k}(0) = {\bar{\bm{\sigma}}_{0,i,k}}, \ 
		\label{eq:r3c4boundary} \\
		&\bm{\sigma}^{[\widetilde d]}_{i,j,k}(\delta T_{i,k}) = \bm{\sigma}^{[\widetilde d]}_{i,j+1,k}(0),\label{eq:r3c4continuity} \\
		&T_{i,k} > 0, \label{eq:r3c4temporal} \\
		&\mathcal{G}_d(\bm{\sigma}(t), ...,\bm{\sigma}^{(s)}(t) , t)\preceq 0,  \forall  d \in  \mathcal{D},  \forall t \in [0, \tau_{k}], \label{eq:r3c4user} \\
		&\forall i \in \{1,2,...,n_k\}, \forall j \in \{1,2,...,M_{i,k}-1\},  \nonumber\\
		&\forall k \in \{1,2,...,K\}, \nonumber
	\end{align}
\end{subequations}
where the definitions of each symbol can be found in our paper.
Eq.(\ref{eq:r3c4boundary}) is the boundary condition where $\bar{\bm{\sigma}}_{0}$ and $\bar{\bm{\sigma}}_{0}$ refer the initial and end states, respectively.  
Eq.(\ref{eq:r3c4continuity}) is the continuity constraint between adjacent polynomials and $\widetilde d$ is the degree of continuity. Based on the proved optimality condition in previous work \cite{wang2021geometrically}, it is required that the minimum control-effort piece-wise polynomial achieves $\widetilde{d}=2s-1$ continuity at waypoints $\bm{p}_{i}$ when the control input dimension is $s$.
For the time positiveness constraint Eq.(\ref{eq:r3c4temporal}), we introduced an unconstrained virtual time $\tau_{i,k}$ and applied a diffeomorphism map to eliminate this constraint:
\begin{equation}
	\begin{aligned}
		T_{i,k} &=\left\{
		\begin{aligned}
			& \frac{1}{2}\tau_{i,k}^2+\tau_{i,k}+1					& \tau_{i,k} > 0 \\
			& \frac{2}{\tau_{i,k}^2-2\tau_{i,k}+2}  					& \tau_{i,k} \leq 0\\
		\end{aligned}
		\right.
	\end{aligned}\label{eq:virtoreal}
\end{equation}
Thanks to the smooth nature of this mapping, the gradient w.r.t $\tau_{i,k}$ can also be analytically computed using chain rules.
We adopt the penalty term  $S_{\Sigma}$ to relax the feasibility constraints Eq. (\ref{eq:r3c4user}):
\begin{equation}
	\begin{aligned}
		S_{\Sigma} = \sum_{d\in \mathcal{D}}w_d
		\sum_{k=1}^{K}&
		\sum_{i=1}^{n_k}
		\sum_{j=1}^{M_{i,k}}
		\int_{t=0}^{\delta T_{i,k}}\mathrm{L}_1(\mathcal{G}_{d}(\bm{\sigma}_{i,j,k}(t), ...,\bm{\sigma}^{(s)}_{i,j,k}(t) , t))dt,\label{eq:smapledpenalty}\\
		\mathrm{L}_1(x) &=\left\{
		\begin{aligned}
			& 0					& x \leq 0 ,\\
			& -\frac{1}{2a_0^3}x^4+\frac{1}{a_0^2}x^3			& 0<x \leq a_0\\
			&x-\frac{a_0}{2}                & a_0<x.
		\end{aligned}
		\right.
	\end{aligned}
\end{equation}
Here, $w_d$ is the penalty weight corresponding to different kinds of constraints. $\mathrm{L}_1(\cdot)$ is  a first-order relaxation function  to guarantee the continuous differentiability and non-negativity of  penalty terms,
and  $a_0=10^{-4}$ is the demarcation point.
In the practical implementation, we discretize the integral in Eq.(\ref{eq:smapledpenalty}) to analytically compute the penalty term. Consequently, the original optimization problem is reformulated into an unconstrained formulation, which can be efficiently solved using the robust L-BFGS algorithm~\cite{liu1989limited}.

\section{The MPC Used in Real-World Experiment}\label{MPC_disscuss}

In this work, the MPC controller serves as a general trajectory tracker to ensure that the robot can effectively follow the planned trajectory. Although we integrated it into our real-world robot system, it is not the primary focus of our contribution, and therefore, we did not allocate significant space for its discussion. We will provide detailed explanations in two aspects:
\begin{table}[]
    \begin{tabular}{|c|c|p{4cm}|c|c|c|p{4cm}|c|}
    \hline
      & \textbf{Symbol} & \textbf{Definition} & \textbf{Index} &   & \textbf{Symbol} & \textbf{Definition} & \textbf{Index} \\ \hline
    1 &  $\bm{\sigma}$            & \multicolumn{1}{|m{4cm}|}{the differential flatness outputs  $\bm\sigma := (\sigma_x, \sigma_y)^T$} & (1)           & 2 & $\bm{c}_{i,j}$   & \multicolumn{1}{|m{4cm}|}{the coefficient matrix of $j$-th piece polynominal in $i$-th segment of the trajectory $\bm{c}_{i,j} \in \mathbb{R}^{2s \times 2}$}               & (1)           \\ \hline
    3 &  $\tau_i$  & \multicolumn{1}{|m{4cm}|}{the start time of the $i$-th segment } & (1) & 4 & $T_i$  & \multicolumn{1}{|m{4cm}|}{the duration of each polynomial in $i$-th segment  }  & (1)           \\ \hline
    5 & $K$           & \multicolumn{1}{|m{4cm}|}{the prediction horizon of MPC  } & (2)     & 6 & $J$ & \multicolumn{1}{|m{4cm}|}{the cost function of MPC}        & (2)         \\ \hline
    7 & $\mathbf{X}_k$    & \multicolumn{1}{|m{4cm}|}{the predictive state at $k$-th step}    & (2)            & 8 & $\mathbf{X}_k^{ref}$    & \multicolumn{1}{|m{4cm}|}{the desired state at $k$-th step  of the trajectory}       & (2)            \\ \hline
    9 & $\bm{U}_k$        & \multicolumn{1}{|m{4cm}|}{the predictive input at $k$-th step}  & (2)            & 10& $\bm{U}_k^{ref}$    & \multicolumn{1}{|m{4cm}|}{the reference input at $k$-th step of the trajectory}    & (2)            \\ \hline
   11&$\bm{Q}_x$&  \multicolumn{1}{|m{4cm}|}{the desired status weights $\bm{Q}_x = diag(w_x,w_y,w_{\theta},w_v)$}& (2)    & 12&$\bm{Q}_u$ & \multicolumn{1}{|m{4cm}|}{Desired input weights $\bm{Q}_u = diag(w_a,w_{steer})$} & (2)           \\ \hline
   13 & $\bm{R}$         & \multicolumn{1}{|m{4cm}|}{the weight of control volume   $\bm{R}=diag(R_{a},R_{steer})$} & (2)& 14 & $\bm{R}_d$ & \multicolumn{1}{|m{4cm}|}{the weight of input smoothing $\bm{R}_d =diag(R_{\Delta a}, R_{\Delta steer})$} & (2) \\ \hline
   15 & $\bm{A}_{k}$ & \multicolumn{1}{|m{4cm}|}{the linearized state transfer matrix at $k$-th step} & (2) & 16 & $\bm{B}_{k}$  & \multicolumn{1}{|m{4cm}|}{the linearized input matrix at $k$-th step} & (2) \\ \hline
   17 & $\bm{U}_{max}$ & \multicolumn{1}{|m{4cm}|}{the maximum input limit $\bm{U}_{max}=diag(a_{max}, \phi_{max})$} & (2) & 18 & $\bm{X}_{max}$ & \multicolumn{1}{|m{4cm}|}{the maximum state limit $\bm{X}_{max}=diag(\infty, \infty,\infty, v_{max})$} & (2) \\ \hline
   19 & $\bm{U}_{dmax}$&\multicolumn{1}{|m{4cm}|}{maximum input difference limit $\bm{U}_{dmax}=diag(\Delta a_{max}, \Delta \phi_{max})$}&(2) & 20 & $L$ &\multicolumn{1}{|m{4cm}|}{the wheelbase of the car-like robot}&(3)\\ \hline
   21 & $\phi$  &  \multicolumn{1}{|m{4cm}|}{the steer input} & (3) &   &  &   &  \\ \hline
    \end{tabular}
    \caption{Symbol definition in MPC discussion.}
    \label{tab:symbol_1}
\end{table}

\begin{enumerate}
    \item The role of MPC in the system of real-world experiments.

    Despite considering the robot's kinematic model and associated constraints in our trajectory planner, practical execution may still result in cumulative errors due to factors such as system delays, ultimately preventing precise trajectory tracking. To address this issue, we introduce an MPC controller between the system actuators and our planner. This feedback controller continuously adjusts the actuator outputs based on the discrepancy between the desired and current states, ensuring robust trajectory following by the robot. Actually, there is no direct coupling between MPC and the proposed planner.

    \item Specific applications of MPC

    Regarding the specific application of MPC, we introduce various mathematical symbols, as shown in Tab. \ref{tab:symbol_1}. It's important to note that we have omitted the subscripts for distinguishing different agents. Each agent receives the trajectory $(\bm{c}, \bm{T})$ and starts analyzing the trajectory from the current time, denoted as the starting time $t_0=0$. At a cumulative time $t$, the differential flatness output $\bm\sigma$ can be expressed as:
    \begin{equation}
        \begin{aligned}
        &\qquad\qquad \bm{\sigma}(t) = \bm{c}^T_{i,j}\bm{\beta}(t), \bm{\beta}(t):=[1,t,\dots,t^5]^T, \\
        & \forall t \in \{t>0: \tau_i+(j-1)T_{i}<t<\tau_i+jT_i,j\in  \{1,2,...,M_i\}  \}
        \end{aligned}
    \end{equation}
    By utilizing the aforementioned expressions, we can obtain the sigma at any given time and its derivatives at various orders. Leveraging the differentiability property, we can further derive the desired body angle $\theta$, velocity $v$, reference input acceleration $a$, and steering angle $\phi$ at any specific time.
    Each execution cycle of the MPC analyzes K timesteps with a time interval of $\Delta t$ between each timestep. Here, we define a state point denoted as  $\bm{X}_k:=\{x_k,y_k,\theta_k,v_k\}$.
    We modify the QP problem formulation for MPC optimization based on the reference \cite{Kong2015KinematicAD}. The cost function can be expressed as:
    \begin{equation}
        \begin{aligned}
            \min_{U_k,\forall k}{}
             J=\sum_{k=1}^{K}&||\bm{X}_k-\bm{X}_k^{ref}||_{\bm{Q}_x}^2+\sum_{k=0}^{K-1}(||\bm{U}_k-\bm{U}_k^{ref}||_{\bm{Q}_u}^2+||\bm{U}_k||_{\bm{R}}^2)&+\sum_{k=1}^{K-1}||\bm{U}_k-\bm{U}_{k-1}||_{\bm{R}_d}^2\\
             \text{s.t.}\
            & \bm{X}_{k+1}=\bm{A}_k \bm{X}_k+\bm{B}_k\bm{U}_k+ \bm{C}_k,& \quad k=0,1,...,K-1,\\
            & ||\bm{U}_k||\le \bm{U}_{max},&\quad k=0,1,...,K-1,\\
            & ||\bm{X}_k||\le \bm{X}_{max},&\quad k=1,2,...,K-1,\\
            & ||\bm{U}_{k}-\bm{U}_{k-1}||\le \bm{U}_{dmax},&\quad k=1,2,...,K-1,\\
        \end{aligned}
    \end{equation}
    In the following section, we will elaborate on the details encountered in the optimization problem. The model employed in this study is the same as described in the paper and can be formulated as:
    \begin{equation}
        \begin{cases}
        \begin{aligned}
        \dot{s}    & = v\cos\theta, \\
        \dot{y}    &= v\sin\theta,          \\
        \dot{\theta} & = \frac{v}{L}\tan\phi, \\
        \dot{v}      & = a,
        \end{aligned}
        \end{cases}
    \end{equation}
    To discretize the model, we can write it in the following form:
    \begin{equation}
        \begin{cases}
            x_{k+1}=x_k-\hat v\cdot \sin\hat\theta\cdot(\theta_k-\hat\theta)\cdot\Delta t+\cos\hat\theta\cdot(v_k-\hat v)\cdot \Delta t+\hat v\cdot\cos\hat\theta\cdot \Delta t,\\
            y_{k+1}=y_k+\hat v\cdot \cos\hat\theta\cdot(\theta_k-\hat\theta)\cdot\Delta t+\sin\hat\theta\cdot(v_k-\hat v)\cdot \Delta t+\hat v\cdot\sin\hat\theta\cdot \Delta t,\\
            v_{k+1}=v_k+(a_k-\hat a)\cdot \Delta t+\hat a\cdot \Delta t,\\
            \theta_{k+1}=\theta_k+(\omega_k-\hat\omega)\cdot \Delta t+\hat\omega\cdot\Delta t.
            \end{cases}
    \end{equation}
    When solving the problem, we linearize each point and can express it as:
    \begin{equation}
        \bm{X}_{k+1}=\bm{A}(\bm{\hat X},\bm{\hat U})\bm{X}_k+\bm{B}(\bm{\hat X},\bm{\hat U})\bm{U}_k+\bm{C}(\bm{\hat X},\bm{\hat U})
    \end{equation}
    Finally, the MPC problem is solved using the OSQP library \cite{osqp}.
\end{enumerate}

\section{Ablation Experiments}\label{ablation}

We have conducted additional experiments on catching and transporting to better demonstrate its effectiveness. Furthermore, we have expanded the comparative analysis and discussion based on the original experiments.

\begin{figure}[H]
	\centering
	{\includegraphics[width=16.0cm]{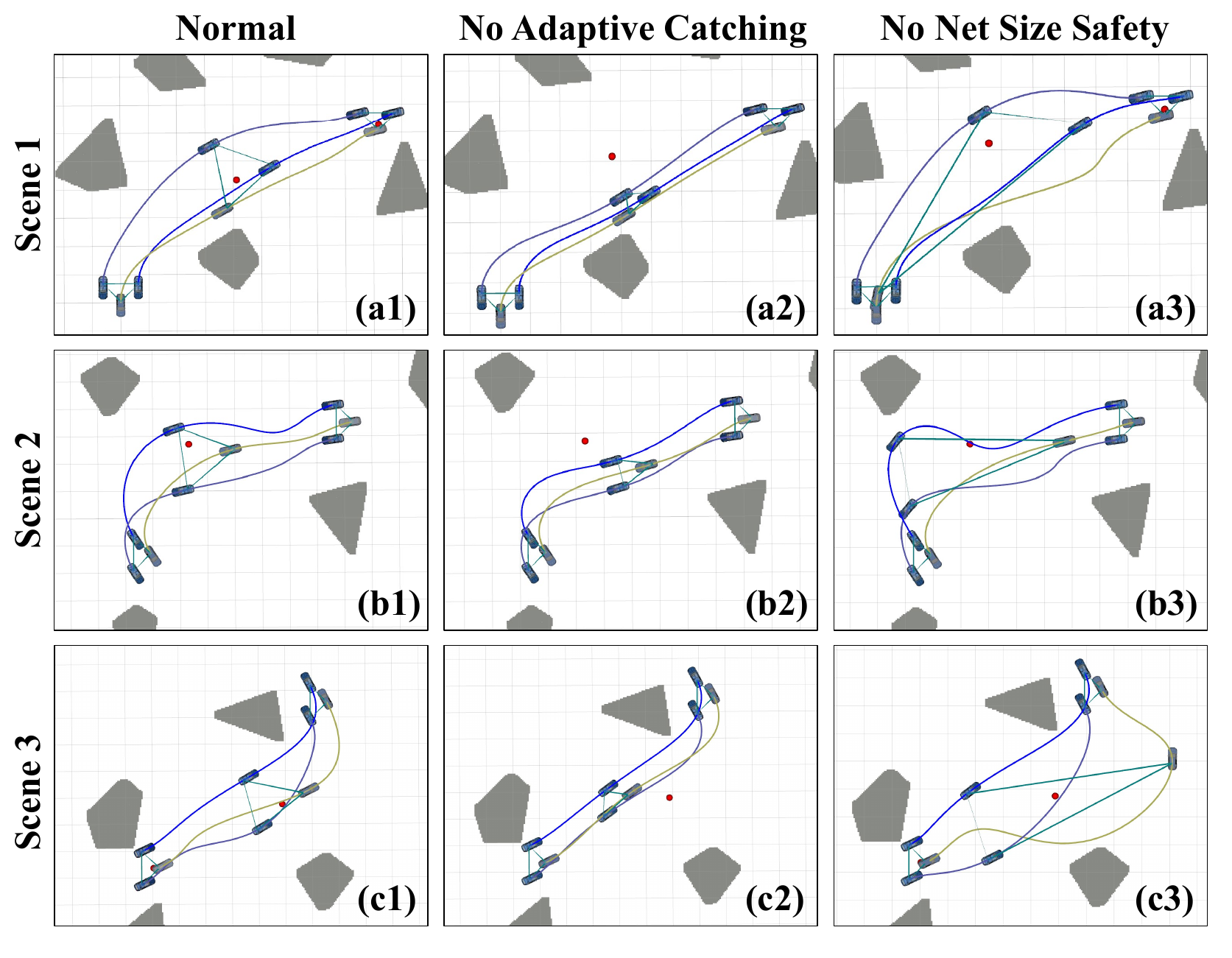}}
	\caption{\label{fig:ablation}In rows labeled 'a', 'b', and 'c', we present three distinct scenarios. In columns '1', '2', and '3', we conduct comparative experiments involving identical initial conditions, identical objectives, and varying functional applications. We display the state of the robot team at the beginning, end of trajectory execution, and when the target is reached. And the target that needs to be caught is visualized as a red sphere.}
\end{figure}

\begin{enumerate}
    \item Efficient and high-quality trajectory generation is the foundation for time-constrained catching and transporting tasks. From the comparative experiments, it is evident that other methods have significant disadvantages in terms of success rate and computation time.  In scenarios where other methods generally exhibit lower trajectory generation efficiency, generating catching and transporting trajectories while considering additional constraints and costs will result in lower success rate and longer computation time. Therefore, for collaborative tasks that require considering more constraints and costs, other methods are less suitable.


    \item To demonstrate the effectiveness of our method in catching and transporting tasks, we have conducted several constraint ablation experiments to showcase the positive impact of our proposed method on collaboration. By removing the net adaptive catching or size safety components, the spatio-temporal trajectories reflect the effectiveness of our proposed method in catching and transporting.

    \begin{enumerate}
        \item As shown in Fig. \ref{fig:ablation}, we compared the trajectories with the adaptive catching component removed. In the landing phase, the robot team did not tend towards the target from the initial positions and did not proactively increase the area of the net to enhance catching probability. This demonstrates the effectiveness of our proposed method in catching.
        \item We compared the trajectories with the net size safety constraint removed. In most cases, it is observed that in order to maximize the probability of successful catching,  agents engaged in aggressive movements, which violated the safety margin of the net size, as illustrated in the third column of  Fig. \ref{fig:ablation}.    
    \end{enumerate}

    \item We have compared our method by incorporating catching and transporting task constraints and costs into the second best-performing method, FOTP, which is suitable for adding collaborative constraints in the benchmark.
\end{enumerate}

\section{Discussion on  $\eta$}\label{eta}

In order to model the vehicle using a differential flatness model, we introduce $\eta$ to represent the forward and backward motion of the robot within a segment of the trajectory, as shown in Eq.(\ref{eq:flatness}).
\begin{equation}\label{eq:flatness}
    \begin{cases}
    \begin{aligned}
    v      & = \eta\sqrt{{\dot{\sigma}_x}^2 + {\dot{\sigma}_y}^2},                                                                                              \\
    \theta & = \arctan2{(\eta{\dot{\sigma}_x}, \eta\dot{\sigma}_y)},                                                                                            \\
    a_t    & =  \eta({\dot{\sigma}_x}\ddot{\sigma}_x + \dot{\sigma}_y\ddot{\sigma}_y) /
    \sqrt{\dot{\sigma}_x^2 + \dot{\sigma}_y^2},                     \\
    a_n    & = \eta(\dot{\sigma}_x\ddot{\sigma}_y - \dot{\sigma}_y\ddot{\sigma}_x)/\sqrt{{\dot{\sigma}_x}^2 + {\dot{\sigma}_y}^2},                              \\
    \phi   & = \arctan\left(\eta(\dot{\sigma}_x\ddot{\sigma}_y - \dot{\sigma}_y\ddot{\sigma}_x)L/({\dot{\sigma}_x}^2 + \dot{\sigma}_y)^{\frac{3}{2}} \right), \\
    \kappa & = \eta(\dot{\sigma}_x\ddot{\sigma}_y - \dot{\sigma}_y\ddot{\sigma}_x)/({\dot{\sigma}_x}^2 + {\dot{\sigma}_y}^2)^{\frac{3}{2}},
    \end{aligned}
    \end{cases}
    \end{equation}
where $\eta$ is a discrete variable with values $\{-1, 1\}$. If used as a decision variable in the optimization problem, it would be challenging to handle due to its strong nonlinearity and discreteness. As we did not optimize the discrete variable $\eta$, it is acknowledged that we sacrifice a certain solution space and optimality. However, this does not imply that the trajectory planning method proposed in this paper loses a significant degree of freedom. This is because of the following reasons:
\begin{enumerate}[label=(\arabic*)]
    \item The determination of $\eta$ is based on the initial guess provided by the Hybrid A* search. If two segments with opposite $\eta$ values are determined, it is highly likely that an Ackerman-based robot will require both forward and backward movements to reach the desired state.
    In such cases, it is appropriate to focus on adjusting the positions and orientations of the gear shifting points generated by forward and backward movements, aiming to achieve smoother trajectories and ensure that the gear shifting points satisfy task constraints, such as net safety.

    \item In segments of the trajectory where the forward and backward motion differs, the gear shifting point, where the transition between the two occurs, divides the trajectory into two parts with opposite $\eta$ values. We incorporate the position and orientation $(x, y, \theta)$ of the common gear shifting point between two different segments of the trajectory into the decision variables. This provides sufficient freedom for trajectory optimization by adjusting the gear shifting point. Experimental verification has shown that in certain scenarios, the initial values provided by the front-end are rough and conservative. The back-end optimization can then adjust the gear shifting point, resulting in a significantly smoother overall trajectory, as illustrated in Fig. \ref{fig:gear_shift}.

    \begin{figure}[H]
        \centering
        {\includegraphics[width=16.0cm]{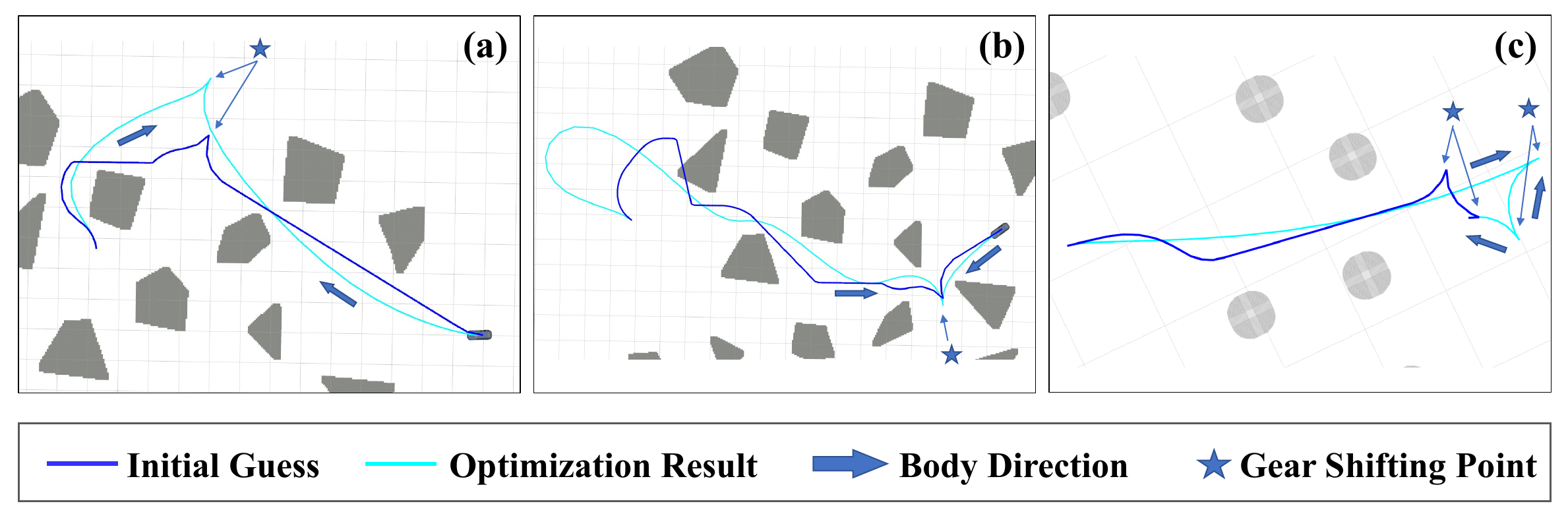}}
        \caption{\label{fig:gear_shift}(a) For initial values that are relatively coarse, the backend performs simultaneous optimization of the entire trajectory, including the gear shifting points. (b) During the optimization of gear shifting points, safety constraints are maintained, ensuring the attainment of a smooth trajectory over the entire segment. (c) In cases involving trajectories with multiple forward and backward transitions, simultaneous optimization of all gear shifting points is achievable.}
    \end{figure}
\end{enumerate}

\section{Forward and Backward Movement of Robotic Team }

In fact, during the implementation of our simulation experiments, there were several setups (including input environment, initial state, and target state) where it was necessary to have both forward and backward movements to reach the target state. However, due to limitations in the length of the paper, we did not emphasize this aspect specifically. 

\begin{figure}[H]
    \centering
    {\includegraphics[width=16.0cm]{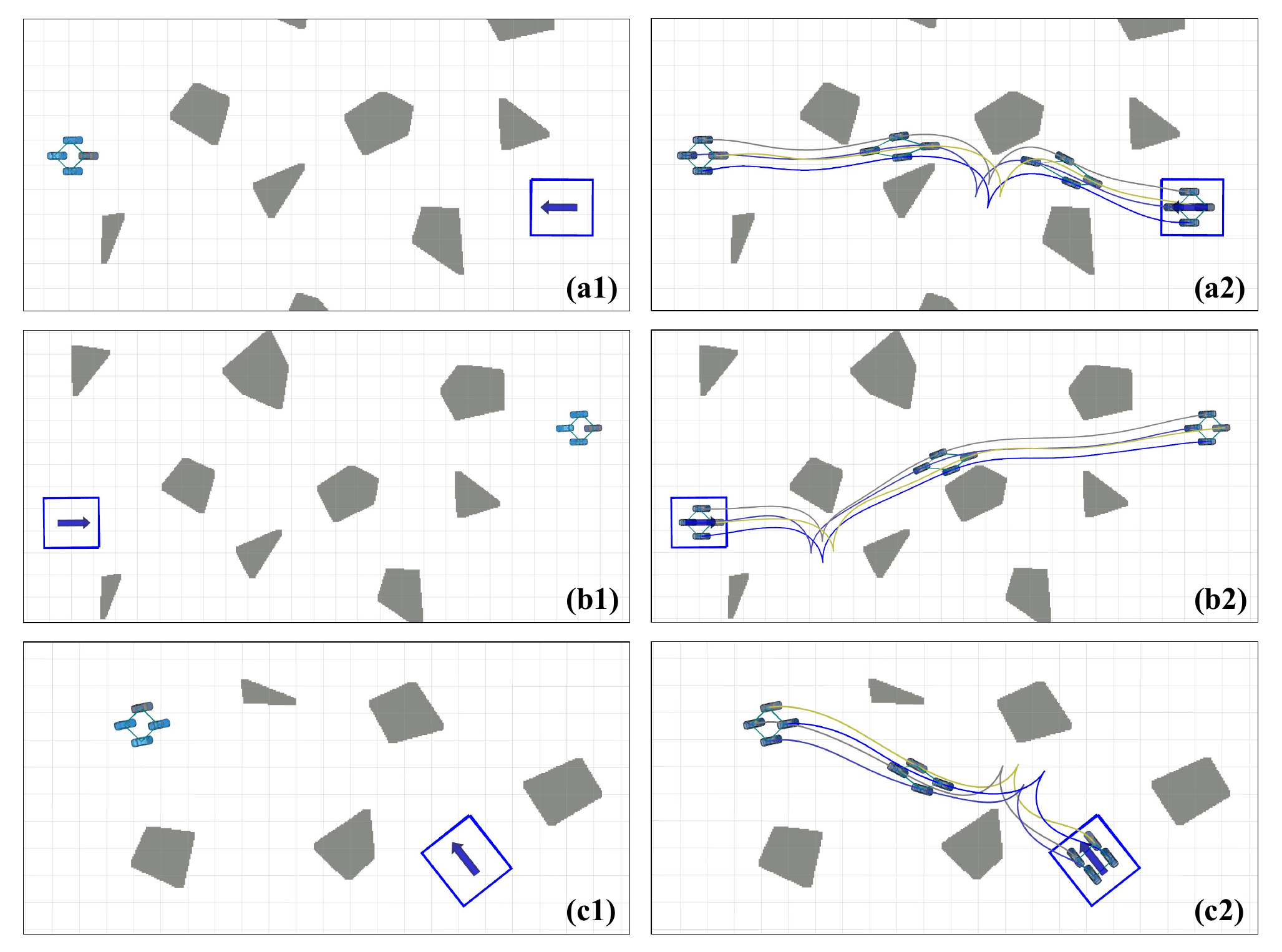}}
    \caption{\label{fig:fbtrans}On the left column, the initial states of robot teams in different environments are displayed, along with the desired states represented by blue arrows. On the right column, the proposed method generates trajectories for robot teams that encompass both forward and backward motion.}
\end{figure}

To address this concern, we have included trajectory planning results showcasing both forward and backward movements in different scenarios for both single agent and multi-agent systems. As shown in Fig. \ref{fig:fbtrans}, the blue trajectory represents the initial guess, the red trajectory depicts the solution result, the green rectangle denotes the initial state, and the purple rectangle represents the target state. We have also included a concise version of these results in the revised version of the manuscript.

\section{Details on Comparative Experiments}\label{detail_com}

We will further supplement the information on the implementation of the comparative experiments and include it in the manuscript. To ensure fairness in our comparative experiments, we conducted the experiments under identical scenarios and inputs.

\begin{enumerate}
    \item To ensure fairness in the comparative experiments, we provided the same initial guess for methods using the Hybird A* algorithm \cite{dolgov2010path}.  For each robot, the algorithm searches for a collision-free path considering the initial and final states of each robot and the environmental obstacles. The search process rejects motion primitives that result in collisions with obstacles and eventually obtains a feasible sequence of states and corresponding inputs. These feasible states and inputs are then used as suitable initial values for each method. It is worth mentioning that CL-CBS \cite{wenCLMAPFMultiAgentPath2022} is a search-based method that obtains collision-free trajectories, so it does not require an initial guess. DMPC \cite{luisOnlineTrajectoryGeneration2020}, on the other hand, continuously solves the optimization problem during planning to generate trajectories for the robot team, and therefore does not rely on an initial guess.
    \item Concerning the cost function, in our comparative experiments, our primary focus lies in assessing the efficiency and quality of each method in generating trajectories for the robot team. This serves as the foundational aspect for achieving efficient catching and transporting. To align with the inherent characteristics of each method and to enhance the likelihood of success, we have employed the original formulations of the cost function as presented in their respective source papers. Broadly, these formulations involve considerations related to trajectory smoothness and total duration. However, each method employs different approaches to achieve smoothness, as indicated in the Tab. \ref{tab:cost_function}.

    \item Regarding the relaxation of final states, we did not incorporate any relaxation of final states in the comparative experiments. This decision was made to ensure fairness in the comparative evaluation. Additionally, the relaxation of final states is mainly introduced to facilitate smoother and safer trajectories after catching the target.
\end{enumerate}

    \begin{table}[]
	\begin{tabular}{|c|c|c|c|}
		\hline
		& \textbf{Method} & \textbf{Cost Funciton} & \textbf{Definition}  \\ \hline
		1 &  FOTP\cite{ouyangFastOptimalTrajectory2022b}  & {$J=T+\mathrm{w} \cdot \sum\limits_{i=1}^{N_{\mathrm{v}}}\left(\sum\limits_{j=0}^H\left(a_{i,j}^2+v_{i,j}^2 \cdot \omega_{i,j}^2\right)  \right)$} & \multicolumn{1}{|m{4cm}|}{ $T$ is the time duration of the trajectories, $w$ is the weight of smooth penalty, $N_v$ is the number of agents, $H$ is the number of planning timesteps, and $a_{i,j},v_{i,j},\omega_{i,j}$ is the acceleration input, velocity, angular  velocity of the $i$-th agent at $j$-th timestep. }                    \\ \hline
		2 &  MNHP  & {$ J=\sum\limits_{i=1}^{N_v}\left(\sum\limits_{j=1}^{H-1} \Delta{u_{i,j}}^{\mathrm{T}} P \Delta u_{i,j}+\sum\limits_{j=1}^H{\bar{z}_{i,j}}^{\mathrm{T}} Q \bar{z}_{i,j}\right)$} &  \multicolumn{1}{|m{4cm}|}{$\Delta u_{i,j} = u_{i,j}-u_{i,j-1}$ is the differences between two successive control inputs, and $\bar{z}_{i,j}$ the deviation between the trajectory and the reference trajectory.  $P \in \mathbb{R}^{2\times 2}$ and  $Q \in \mathbb{R}^{3 \times3}$ are two positive definite weighting matrices.}                  \\ \hline
		3 &  DMPC  &  $J=\sum\limits_{j=H-\kappa}^H q_k\left\|\hat{\mathbf{z}}_i\left[j \mid j_t\right]-\mathbf{z}_{d, i}\right\|_2^2 + \sum\limits_{c=0}^r \alpha_c \int_0^{t_h}\left\|\frac{d^c}{d t^c} \hat{\mathbf{u}}_i(t)\right\|_2^2 d t$ & \multicolumn{1}{|m{4cm}|}{The first term aim to minimize the sum of errors between the positions and the goal state $\mathbf{z}_{d,i}$ at the last $\kappa < H$ timesteps. The second term aim to minimize a  weighted combination of the sum of squared derivatives. }                     \\ \hline
		4 &  SCP  & {$J=\sum\limits_{i=1}^{N_v} \sum\limits_{q=1}^{N_v} \sum\limits_{j=1}^H w_{i,q} a_{i,j}^T a_{q,j}$} & \multicolumn{1}{|m{4cm}|}{$w_{i,j} = 1$ when $i = j$ and $w_{i,j} = 0$ otherwise, which tends to induce smooth trajectories with low curvatures.}                    \\ \hline
	\end{tabular}
	\caption{The cost functions of various methods being compared share a common set of symbol definitions.}
	\label{tab:cost_function}
\end{table}

\section{More Comparative Experiments for FOTP}\label{R3C6}

It is worth mentioning that we conducted several comparative experiments for FOTP \cite{ouyangFastOptimalTrajectory2022b} under different time weightings. This is because there exists a trade-off relationship between trajectory duration and smoothness. As detailed in Section Sec. \ref{detail_com}, the cost function of FOTP can be expressed as:
\begin{equation}
    J=T+\mathrm{w} \cdot \sum\limits_{i=1}^{N_{\mathrm{v}}}\left(\sum\limits_{j=0}^H\left(a_{i,j}^2+v_{i,j}^2 \cdot \omega_{i,j}^2\right)  \right)
\end{equation}
, where $w$ represents the weight assigned to the time duration. In our previous version of the manuscript, we set $w=0.1$ following the original method.  However, FOTP achieves significantly longer trajectory durations compared to other methods in exchange for lower acceleration costs.  Consequently, to understand how the trajectory duration of FOTP compares to other methods while maintaining a similar level of smoothness, we conducted comparative experiments with $w= 0.01$ and $w=0.001$, under the same environment and initial/final states as the previous version. The results are shown in Tab. \ref{tab:FOTP_data}. In order to control the method at the same TRAVEL time level, we adopt the result of $w=0.001$.

\begin{table}[H]
    \renewcommand\arraystretch{1.1}
    \begin{tabular}{|l|c|ccc|c|c|}
    \hline
    \multicolumn{1}{|c|}{\multirow{2}{*}{Method}} &
      \multirow{2}{*}{\begin{tabular}[c]{@{}c@{}}Success \\ Rate\end{tabular}} &
      \multicolumn{3}{c|}{Time(s)} &
      \multirow{2}{*}{\begin{tabular}[c]{@{}c@{}}Average \\ Travel\\ Distance($m$)\end{tabular}} &
      \multirow{2}{*}{\begin{tabular}[c]{@{}c@{}}Average \\ Acceration\\ Cost($m^2s^{-3}$)\end{tabular}} \\ \cline{3-5}
    \multicolumn{1}{|c|}{} &
       &
      \multicolumn{1}{l|}{Computation} &
      \multicolumn{1}{c|}{\begin{tabular}[c]{@{}c@{}}Mean \\ Travel\end{tabular}} &
      \begin{tabular}[c]{@{}c@{}}Longest \\ Travel\end{tabular} &
       &
       \\ \hline
    \begin{tabular}[c]{@{}l@{}}Proposed, N = 8\\ Proposed, N = 14\\ Proposed, N = 20\\ Proposed, N = 26\end{tabular} &
      \begin{tabular}[c]{@{}c@{}}100\%\\ 100\%\\ 100\%\\ 100\%\end{tabular} &
      \multicolumn{1}{c|}{\begin{tabular}[c]{@{}c@{}}0.280\\ 0.707\\ 1.186\\ 1.900\end{tabular}} &
      \multicolumn{1}{c|}{\begin{tabular}[c]{@{}c@{}}5.956\\ 6.067\\ 6.328\\ 6.387\end{tabular}} &
      \begin{tabular}[c]{@{}c@{}}6.918\\ 6.841\\ 7.151\\ 8.216\end{tabular} &
      \begin{tabular}[c]{@{}c@{}}84.461\\ 145.002\\ 207.469\\ 271.725\end{tabular} &
      \begin{tabular}[c]{@{}c@{}}244.925\\ 205.204\\ 196.532\\ 218.752\end{tabular} \\ \hline
    \begin{tabular}[c]{@{}l@{}}$w=0.001$\\ FOTP, N = 8\\ FOTP, N = 14\\ FOTP, N = 20\\ FOTP, N = 26\end{tabular} &
      \begin{tabular}[c]{@{}c@{}} \\ 100\%\\ 100\%\\ 100\%\\ 100\%\end{tabular} &
      \multicolumn{1}{c|}{\begin{tabular}[c]{@{}c@{}} \\ 4.897\\ 37.801\\ 147.214\\ 376.645\end{tabular}} &
      \multicolumn{1}{c|}{\begin{tabular}[c]{@{}c@{}} \\ 6.169\\ 6.957\\ 7.835\\ 8.888\end{tabular}} &
      \begin{tabular}[c]{@{}c@{}} \\ 6.169\\ 6.957\\ 7.835\\ 8.888\end{tabular} &
      \begin{tabular}[c]{@{}c@{}} \\ 81.462\\ 149.151\\ 222.223\\ 297.745\end{tabular} &
      \begin{tabular}[c]{@{}c@{}} \\ 286.648\\ 245.775\\ 204.270\\ 154.748\end{tabular} \\ \hline
    \begin{tabular}[c]{@{}l@{}}$w=0.01$\\ FOTP, N = 8\\ FOTP, N = 14\\ FOTP, N = 20\\ FOTP, N = 26\end{tabular} &
      \begin{tabular}[c]{@{}c@{}} \\ 100\%\\ 100\%\\ 100\%\\ 100\%\end{tabular} &
      \multicolumn{1}{c|}{\begin{tabular}[c]{@{}c@{}}  \\ 4.970\\ 69.084\\ 159.535\\ 384.098\end{tabular}} &
      \multicolumn{1}{c|}{\begin{tabular}[c]{@{}c@{}} \\ 8.492\\ 9.780\\ 10.847\\ 12.100\end{tabular}} &
      \begin{tabular}[c]{@{}c@{}} \\ 8.492\\ 9.780\\ 10.847\\ 12.100\end{tabular} &
      \begin{tabular}[c]{@{}c@{}} \\ 81.545\\ 146.804\\ 214.560\\ 287.217\end{tabular} &
      \begin{tabular}[c]{@{}c@{}} \\ 109.782\\ 81.110\\ 65.331\\ 57.885\end{tabular} \\ \hline
    \begin{tabular}[c]{@{}l@{}}$w=0.1$\\ FOTP, N = 8\\ FOTP, N = 14\\ FOTP, N = 20\\ FOTP, N = 26\end{tabular} &
      \begin{tabular}[c]{@{}c@{}} \\ 100\%\\ 100\%\\ 100\%\\ 100\%\end{tabular} &
      \multicolumn{1}{c|}{\begin{tabular}[c]{@{}c@{}} \\ 5.690\\ 50.111\\ 135.057\\ 344.703\end{tabular}} &
      \multicolumn{1}{c|}{\begin{tabular}[c]{@{}c@{}} \\ 13.789\\ 15.254\\ 17.028\\ 18.332\end{tabular}} &
      \begin{tabular}[c]{@{}c@{}} \\ 13.789\\ 15.254\\ 17.028\\ 18.332\end{tabular} &
      \begin{tabular}[c]{@{}c@{}} \\ 82.561\\ 144.233\\ 209.591\\ 277.040\end{tabular} &
      \begin{tabular}[c]{@{}c@{}} \\ 28.376\\ 20.179\\ 16.890\\ 14.209\end{tabular} \\ \hline
    \end{tabular}
    \caption{The experimental data of FOTP under different time weight $w$}
    \label{tab:FOTP_data}
    \end{table}

\section{Discussion on The Local Minimum}\label{sec:local_minimum}

The objective of the trajectory optimization is to solve a multi-stage Linear Quadratic Minimum Time (LQMT) problem \cite{wang2021geometrically}, which is non-convex and non-linear.
Furthermore, when it comes to the whole-body inter-agent avoidance and adaptive catching, there exist highly nonlinear constraints and costs. Therefore, we cannot guarantee that gradient descent methods will attain the globally optimal solution in optimization.

Local minimum refers to the high-quality trajectory of a robot team obtained through optimization within the topological environment based on the current initial input values. Our experiments demonstrate that the optimized trajectory is smoother compared to the initial values and eliminates infeasible points on the initial trajectory, as shown in Fig. \ref{fig:ablation} and Fig. \ref{fig:gear_shift}. However, we acknowledge that, in the context of nonlinear optimization, the initial values play a crucial role as they affect the quality of the final solution and the solution time.In practical engineering applications, we employ the widely-used and efficient Hybrid A* algorithm \cite{dolgov2010path} to obtain the initial values, which can generate a curvature-constrained and collision-free path. Nevertheless, it is possible that the initial values may not completely satisfy all constraints or achieve successful capture at certain points. However, extensive experimentation has demonstrated that thanks to the powerful optimization capability of our backend planning algorithm, the optimized trajectory can often accomplish the tasks of catching and transporting while ensuring safety and dynamic feasibility.

\section{Discussion on The Approximations Applied}\label{sec:approximaiton}

We introduce the smoothing approximations in order to achieve reasonable gradient descent directions throughout the entire optimization process, ensuring that the solver efficiently and accurately converges to local minimum points. In practical engineering applications, we believe that the errors caused by this smoothing term can be almost negligible in determining the feasibility of finding a solution, which is further validated by the high success rate observed in experiments. However, we acknowledge that in principle, smoothing can sometimes lead to planning failures in situations where success should have been achieved. Nevertheless, we consider smoothing to be unavoidable because there exist non-differentiable points in the signed distance between two convex polygons in physical terms. Additionally, despite the introduction of smoothing, our full-shape collision avoidance method is more precise compared to particle-based models.
We provide explanations regarding the details of the approximations, the algorithm itself, and the application specifics:

\begin{enumerate}[label=(\arabic*)]
    \item The inclusion of smoothing terms aims to obtain better gradients within a small neighborhood around the constraint edge when performing gradient descent. This approach prevents the occurrence of abnormal gradients that may result in the failure of gradient descent in that region. Accordingly, the parameters related to the smoothing process are set to very small values and only take effect within a tiny neighborhood of critical points, such as $\alpha$ in LSE and $\epsilon \in \mathbb{R}^+$  in $L(x)$ of adaptive catching. This effectively mitigates the influence of the approximations on the optimization process.
    \item In fact, in our simulation or comparative experiments, we have not encountered instances where the smoothing leads to non-existent solutions. On the contrary, our method has demonstrated a higher success rate compared to conservative approaches. Although we employ smoothing, the effective solution space for obstacle avoidance, considering the whole-body aspect, remains larger compared to conservative particle-based models.
    \item To tackle safety issues arising from infeasible solutions, there are some details:
    \begin{enumerate}
        \item With the cost $\Psi$ illustrated in IV.C, adaptive catching encourages the active expansion of the net when catching target and bringing the center of the net closer to the target.   This provides a margin for catching, and has been observed to significantly improve the success rate of catching in simulations.
        \item  Assigning greater weights to environmental obstacle avoidance and inter-agent collision avoidance during the solution process, prioritizing agent safety.
        \item Setting conservative values for physical parameters related to net size safety and net topological safety, allowing for a margin of safety in net safety.
    \end{enumerate}
\end{enumerate}

In summary, these three points mitigate the impact of approximations on the solution space of our method. While it is true that we may encounter a reduction in solution space theoretically, our approach offers greater advantages in terms of success rate and computation time compared to conservative methods. Furthermore, through the design of our method itself, we effectively achieve coordinated catching and transporting in practical applications while balancing computational efficiency, success rate, and solution quality.

\bibliography{references}

\begin{thebibliography}{1}
\providecommand{\url}[1]{#1}
\csname url@samestyle\endcsname
\providecommand{\newblock}{\relax}
\providecommand{\bibinfo}[2]{#2}
\providecommand{\BIBentrySTDinterwordspacing}{\spaceskip=0pt\relax}
\providecommand{\BIBentryALTinterwordstretchfactor}{4}
\providecommand{\BIBentryALTinterwordspacing}{\spaceskip=\fontdimen2\font plus
\BIBentryALTinterwordstretchfactor\fontdimen3\font minus \fontdimen4\font\relax}
\providecommand{\BIBforeignlanguage}[2]{{%
\expandafter\ifx\csname l@#1\endcsname\relax
\typeout{** WARNING: IEEEtran.bst: No hyphenation pattern has been}%
\typeout{** loaded for the language `#1'. Using the pattern for}%
\typeout{** the default language instead.}%
\else
\language=\csname l@#1\endcsname
\fi
#2}}
\providecommand{\BIBdecl}{\relax}
\BIBdecl

\bibitem{wang2021geometrically}
Z.~Wang, X.~Zhou, C.~Xu, and F.~Gao, ``Geometrically constrained trajectory optimization for multicopters,'' \emph{IEEE Transactions on Robotics}, pp. 1--10, 2022.

\bibitem{liu1989limited}
D.~C. Liu and J.~Nocedal, ``On the limited memory bfgs method for large scale optimization,'' \emph{Mathematical programming}, vol.~45, no. 1-3, pp. 503--528, 1989.

\bibitem{Kong2015KinematicAD}
J.~Kong, M.~Pfeiffer, G.~Schildbach, and F.~Borrelli, ``{Kinematic and dynamic vehicle models} for {autonomous driving control design},'' \emph{2015 IEEE Intelligent Vehicles Symposium (IV)}, pp. 1094--1099, 2015.

\bibitem{osqp}
\BIBentryALTinterwordspacing
B.~Stellato, G.~Banjac, P.~Goulart, A.~Bemporad, and S.~Boyd, ``{OSQP}: an operator splitting solver for quadratic programs,'' \emph{Mathematical Programming Computation}, vol.~12, no.~4, pp. 637--672, 2020. [Online]. Available: \url{https://doi.org/10.1007/s12532-020-00179-2}
\BIBentrySTDinterwordspacing

\bibitem{dolgov2010path}
D.~Dolgov, S.~Thrun, M.~Montemerlo, and J.~Diebel, ``Path planning for autonomous vehicles in unknown semi-structured environments,'' \emph{The International Journal of Robotics Research}, vol.~29, no.~5, pp. 485--501, 2010.

\bibitem{wenCLMAPFMultiAgentPath2022}
L.~Wen, Z.~Zhang, Z.~Chen, X.~Zhao, and Y.~Liu, ``{{CL-MAPF}}: {{Multi-Agent Path Finding}} for {{Car-Like Robots}} with {{Kinematic}} and {{Spatiotemporal Constraints}},'' \emph{Robotics and Autonomous Systems}, vol. 150, p. 103997, Apr. 2022.

\bibitem{luisOnlineTrajectoryGeneration2020}
C.~E. Luis, M.~Vukosavljev, and A.~P. Schoellig, ``Online {{Trajectory Generation With Distributed Model Predictive Control}} for {{Multi-Robot Motion Planning}},'' \emph{IEEE Robotics and Automation Letters}, vol.~5, no.~2, pp. 604--611, Apr. 2020.

\bibitem{ouyangFastOptimalTrajectory2022b}
Y.~Ouyang, B.~Li, Y.~Zhang, T.~Acarman, Y.~Guo, and T.~Zhang, ``Fast and {{Optimal Trajectory Planning}} for {{Multiple Vehicles}} in a {{Nonconvex}} and {{Cluttered Environment}}: {{Benchmarks}}, {{Methodology}}, and {{Experiments}},'' in \emph{{{International Conference}} on {{Robotics}} and {{Automation}} ({{ICRA}})}, May 2022, pp. 10\,746--10\,752.

\end{thebibliography}
\end{document}